\documentclass{article}       
\usepackage{graphicx}
\usepackage{amssymb}
\usepackage{amsmath}
\usepackage{epstopdf}
\usepackage[tight,footnotesize]{subfigure}
\usepackage{stfloats}
\usepackage{microtype}
\usepackage{units}
\usepackage{algorithmic}
\usepackage{algorithm}
\usepackage{url}
\usepackage{bm}
\usepackage{multirow}

\DeclareMathOperator*{\argmax}{arg\,max}
\usepackage{float}
\usepackage{adjustbox}
\graphicspath{{figures/}}

\begin{document}
\title{The Heterogeneous Ensembles of Standard Classification Algorithms (HESCA): the Whole is Greater than the Sum of its Parts.}

\author{James Large, Jason Lines and Anthony Bagnall\\
 School of Computing Sciences \\
              University of East Anglia \\
              United Kingdom\\
}
\maketitle

\begin{abstract}
Building classification models is an intrinsically practical exercise that requires many design decisions prior to deployment. We aim to provide some guidance in this decision making process. Specifically, given a classification problem with real valued attributes, we consider which classifier or family of classifiers should one use. Strong contenders are tree based homogeneous ensembles, support vector machines or deep neural networks. All three families of model could claim to be state-of-the-art, and yet it is not clear when one is preferable to the others. Our extensive experiments with over 200 data sets from two distinct archives demonstrate that, rather than choose a single family and expend computing resources on optimising that model, it is significantly better to build simpler versions of classifiers from each family and ensemble. We show that the Heterogeneous Ensembles of Standard Classification Algorithms (HESCA), which ensembles based on error estimates formed on the train data, is significantly better (in terms of error, balanced error, negative log likelihood and area under the ROC curve) than its individual components, picking the component that is best on train data, and a support vector machine tuned over 1089 different parameter configurations. We demonstrate HESCA+, which contains a deep neural network, a support vector machine and two decision tree forests, is significantly better than its components, picking the best component, and HESCA. We analyse the results further and find that HESCA and HESCA+ are of particular value when the train set size is relatively small and the problem has multiple classes. HESCA is a fast approach that is, on average, as good as state-of-the-art classifiers, whereas HESCA+ is significantly better than average and represents a strong benchmark for future research.

\end{abstract}

\section{Introduction}

Investigation into the properties and characteristics of classification algorithms forms a significant component of all research in machine learning. Broadly speaking, there are three families of algorithm that could claim to be state-of-the-art: support vector machines; multilayer perceptrons/deep learning; and tree based ensembles. Nevertheless, there are still good reasons, such as scalability and interpretability, to use simpler classifiers such as decision trees and nearest neighbour classifiers. Thousands of publications have considered variants of these algorithms on a huge range of problems and scenarios. Sophisticated theories into performance under idealised conditions have been developed, and tailored models for specific domains have achieved impressive results. However, data mining is an intrinsically practical exercise, and our interest is in answering the following question: if we have a new classification problem or set of problems, what family of models should we use given my computational constraints? This interest has arisen from our work in the domain of time series classification~\cite{bagnall16bakeoff}, and through working with many industrial partners, but we cannot find an acceptable answer in the literature. The comparative studies of classifiers give some indication (for example~\cite{delgado14hundreds}), but most people make the decision for pragmatic or dogmatic reasons. We touch on the broad (and highly contentious) issue of which classifier is better on average over standard problems, but do not claim to offer a definitive answer.  Instead, our key hypothesis is that, in the absence of specific domain knowledge, it is in fact better to ensemble classifiers from different families rather than intensify computational efforts into optimising a specific type. Our primary contribution is to demonstrate that a simple ensembling scheme can make small sets of different classification algorithms better. It could be argued that this is hardly a novel observation. It is widely known and accepted that ensembling improves weak classifiers. However, the vast majority of research into ensembles has focused on combining identical algorithms. We do not believe that most practitioners are aware that, on average, a significant improvement in accuracy can be achieved through the simple expedient of combining algorithms commonly available in many software packages, even if there is no significant difference between the constituents. The embarrassingly parallel nature of simple ensembling means that the actual ensembling can be done independently of individual model building. Our contribution is to address a number of questions relating to simple, general purpose ensembling.

\begin{enumerate}
\item Does ensembling classifiers that are not on average significantly different significantly improve overall performance?
\item How well can we determine which classifier to use with just the train data and is this better than ensembling?
\item Is there any significant difference between alternative ways of ensembling?
\item Is it better to tune a single classifier than to ensemble minimally tuned classifiers?
\item Can we use the ensemble to gain insights into the performance of tuned base classifiers?
\end{enumerate}

To begin answering these questions, we have to clarify what we mean when we say one classifier is better than another. We compare classifiers on unseen data based on the quality of the decision rule (using classification error and balanced classifier error to account for class imbalance) the ability to rank cases (with the area under the receiver operator curve) and the probability estimates (using negative log likelihood). We also assess how good a classifier is at predicting the test error from cross validation on the train data.  To control for one source of variation, we restrict our attention to data with continuous attributes only. We compare over multiple resamples on a range of data using standard statistical parametric and non-parametric tests. We perform this evaluation using two sets of public repository data sets. We use 121 data derived from the UCI archive in~\cite{delgado14hundreds} and 85 time series data from the UCR-UEA archive~\cite{TSCWeb}. We compare a range of weighting schemes that have been proposed in the literature and conclude that the simple mechanism of weighting based on estimates of error derived on the train data is as good an approach to weighting as any other. We conclude that, on average, choosing a classifier based on estimates of error from the train set is significantly worse than using the simple classifier weighting scheme we call the Heterogeneous Ensembles of Standard Classifiers (HESCA), which also significantly improves the constituents on average. These results hold on both sets of data sets.  We compare two versions of HESCA to a support vector machine with a tuned spherical Radial Basis Function, and find HESCA to be significantly better.
We further investigate whether the characteristics of the data are indicative of whether selecting a classifier is inferior to ensembling and find, unsurprisingly, ensembling is better when there are fewer training cases, but overall there no clear pattern.

Our conclusion and recommendation to practitioners is that if the computing resources are available, it is, on average, better to ensemble strong classifiers with a weighting scheme based on cross validated estimates of error such as HESCA and that is a sensible starting point for any problem with real valued attributes.

The remainder of this paper is structured as follows. Section~\ref{background} provides an overview of recent experimental comparisons of classifiers, a description of the statistics we measure, tests we use and some basic background into ensemble methods.
Section~\ref{hesca} describes the HESCA classifier and motivates the design decisions made in its definition. The results on UCI data sets are presented in Section~\ref{results}. We delve deeper into the UCI results in Section~\ref{analysis}. We then examine whether our results are reproducible on a completely different set of data by experimenting with the UCR-UEA time series classification data sets in Section~\ref{TSC}. Finally, we conclude in Section~\ref{conc}.

\section{Background}
\label{background}
\subsection{Comparing Classifiers}

The UCI dataset archive\footnote{http://archive.ics.uci.edu/ml/index.php} is widely used in the machine learning and data mining literature, with subsets of the wide range of different dataset types used to evaluate proposed algorithms. An extensive evaluation of 179 classifiers on 121 datasets from the UCI archive, including different implementations of notionally the same classifier, was performed by \cite{delgado14hundreds}. The datasets chosen were selected or converted to be real-valued only.

Overall, they found that the Random Forest (RandF) algorithms maintained the highest average ranking, with Support Vector Machines (SVM) and Neural Networks achieving comparable performance. There was no algorithm significantly better than all others on average. Although it has since been identified that the overlap between validation and test data sets may have introduced bias~\cite{Wainberg16randomforests}, these results mirror our own experience with these classifiers.

The UCR-UEA archive is a continually growing collection of real valued time series classification (TSC) datasets\footnote{http://www.timeseriesclassification.com}. A recent study~\cite{bagnall16bakeoff} implemented 18 state-of-the-art TSC classifiers within a common framework and evaluated them on 85 datasets in the archive. The best performing algorithm, the Collective of Transformation-based Ensembles (COTE), was a heterogeneous ensemble of strong classifiers. These results were our primary motivation for further exploring heterogeneous ensembles for classification problems in general.

While perhaps not feasible or even necessary for every new algorithm that appears, large scale experiments such as these provide a key foundation for comparative evaluation in new literature. They aid clarity and ease of assessment for claims made for a new classifier, be that general improvement or improvement within some particular domain.

\subsection{Performance Statistics}
\label{stats}
A data set $D$ of size $n$ is a set of attribute vectors with an associated observation of a class variable (the response), $D=\{(\bm{x_1},y_1),\ldots,(\bm{x_n},y_n)\}$, where the class variable has $c$ possible values, $y \in \{1,\ldots,c\}$. We assume we can iterate over the elements $\bm{x}$ or $y$ in $D$    by index $i$. Suppose we have a classifier, $M$, constructed on train data $D_{r}$, which we evaluate on a test data set $D_{e}$. To avoid any ambiguity, we stress that all model selection, parameter tuning and/or model fitting that may occur are conducted on the train set, which may or may not require nested cross validation. The final resulting classifier, $M$, is built once on $D_r$ and applied only once to any test set $D_{e}$.

A classifier is a mapping from the space of possible attribute vectors to the space of possible probability distributions over the $c$ valid values of the class variable, $M(\bm{x})=\hat{\bm{p}}$, where $\hat{\bm{p}}=\{\hat{p}(y=1|\bm{x}),\ldots,\hat{p}(y=c|\bm{x})\}$. Given $\hat{\bf{p}}$, the estimate of the response is simply the value with the maximum probability, i.e.

$$\hat{y}=\argmax_{j=1,\ldots,c} \hat{p}(j).$$
A correctness function $f(y,\hat{y})$ returns 1 if the prediction is correct, zero otherwise,
\[    f(y,\hat{y})=
\begin{cases}
    1,& \text{if } y=\hat{y}\\
    0,              & \text{otherwise}
\end{cases}
\]
The test set error is simply the proportion of incorrect predictions
\begin{equation}
e(D_{e}|M,D_{r})=1-\frac{\sum_{y_i \in D_{e}} f(y_i,\hat{y_i})}{|D_{e}|}.
\label{error}
\end{equation}
On some occasions in the results we refer to the accuracy (one minus the error) for clarity. To compensate for class imbalance, we also examine the balanced error rate. If we define the proportion correct in the test set for each class $j$ as
$$s_j=\frac{\sum_{y_i \in D_{e},y_i=j}f(y_i,\hat{y_i})}{\sum_{y_i \in D_{e}}f(y_i,j)},$$
and denote $r_j$ as the proportion of class $j$ in the train data, then the balanced error is
\begin{equation}
e_b(D_{e}|M,D_{r})=\sum_{j=1}^c r_j \cdot s_j.
\label{balancederror}
\end{equation}
The likelihood is the probability of having observed the test data given our classifier, i.e.
$$L(D_{e}|M,D_{r})=\prod_{\bm{x_i} \in D_{e}} \hat{p}(y_i|\bm{x}_i,M).$$
The likelihood will be zero if the classifier predicts zero probability for the true class for any test instance. This limits the usefulness of the statistic, as it can significantly skew the results. For this reason we normalise all probability estimates when calculating the likelihood so that the minimum probability for any one class is 0.01. To make comparison with error more meaningful, we assess classifiers with the negative log likelihood (NLL),
\begin{equation}
l(D_{e}|M,D_{r})=\sum_{x_i \in D_{e}} \log_2(\hat{p}(y_i|\bm{x}_i,M)).
\label{nll}
\end{equation}
The fourth statistic is the area under the receiver operator characteristic curve (AUROC). AUROC is best defined where one class is considered a `success'. Suppose we designate $y=1$ a success and all other outcomes a failure. The classifier predictions of the probability of a success for the $n$ instances in $D_{e}$ as $\hat{p}=\{\hat{p}_1,\ldots,\hat{p}_n\}$. Observed values of the response are $\{y_1,\ldots,y_n\}$. The AUROC is based on the order statistics. We let $\hat{p}_{(i)}$ denote the $i^{th}$ order statistic (in descending order) and $y_{(i)}$ the observed value of the response associated with probability estimate $\hat{p}_{(i)}$. These values are then used as classification functions $d(i,j)$, where 1 is a success and 0 a failure,

\[
\hat{y}_{(j)}=d(i,j)=\begin{cases}
1, & \text{if  } j \leq i\\
0, & \text{otherwise}
\end{cases}
\]
The ROC curve is a series of $n$ points representing the false positive rate (the proportion of failures classified as a success) on the x-axis and the true positive rate (proportion of actual successes classified as a success) on the y-axis each associated with a decision boundary.
So, for example, if there are $a$ positive cases and $b$ negative ($a+b=n$), then, for any point $i$, the decision boundary is to classify as positive only those with probability greater than or equal to $\hat{p}_{(i)}$. The true positive rate is given by $$tpr_i=\frac{\sum_{j=1}^i f(y_{(j)},d(i,j))}{a},$$ and  the false positive rate is  $$fpr_i=\frac{\sum_{j=1}^i (1-f(y_{(j)},d(i,j)))}{b}.$$
Given a list of $n$ points $$t=<(fpr_1,tpr_1),\ldots, (fpr_n,tpr_n)>$$ from the $n$ decision boundaries, the ROC curve is a subset of this list consisting of pairs with unique point $fpr$ values. If there are duplicate $fpr$ values in $t$, the one with the maximum $tpr$ is selected for the $ROC$. (0,0) is inserted at the beginning and(1,1) at the end. Given then a ROC curve

$$ROC=<(a_1,b_1),\ldots,(a_k,b_k)>$$
If class $s$ is judged success, AUROC is defined as
\begin{equation*}
AUROC_s(D_{e}|M,D_{r})=\sum_{i=2}^{k} a_i\cdot (b_{i+1}-b_i)
\label{auroc}
\end{equation*}
For problems with two classes, we treat the minority class as a success. For multiclass problems, we calculate the AUROC for each class and weight it by the class frequency in the train data, as recommended in~\cite{provost03pet},

\begin{equation}
AUROC(D_{e}|M,D_{r})=\sum_{i=1}^{c} w_i \cdot AUROC_i(D_{e}|M,D_{r})
\label{mcauroc}
\end{equation}
The final statistic we use is the difference between estimated test set error, found on the train set, and true test set error. To estimate test accuracy from the train data we cross validate. We perform all model selection being separately on each train fold within the cross validation and evaluate only once on the test fold, using the statistics defined above. 

\subsection{Tests of Difference Between Classifiers}

For any one data set we perform a number of stratified resamples into train and test sets. We always compare classifiers on the same resamples, and these can be exactly reproduced with the published code. This means we can compare two classifiers with paired two sample tests, such as Wilcoxon sign rank test. For comparing two classifiers on multiple datasets we  compare either the number of data sets where there is a significant difference over resamples, or we can do a pairwise comparison of the average errors over all folds.

For comparing multiple classifiers on multiple data sets, we follow the recommendation of Dem\v{s}ar~\cite{demsar06comparisons} and use the Friedmann test to determine if there were any statistically significant differences in the rankings of the classifiers.
However, following recent recommendations in \cite{benavoli16pairwise} and \cite{garcia08pairwise}, we have abandoned the Nemenyi post-hoc test originally used by~\cite{demsar06comparisons} to form cliques (groups of classifiers within which there is no significant difference in ranks). Instead, we compare all classifiers with pairwise Wilcoxon
signed rank tests, and form cliques using the Holm correction (which adjusts family-wise error less conservatively than a Bonferonni adjustment).

\subsection{Ensemble Methods}
\label{enembles}
The key concept in ensemble design is the requirement to inject diversity into the ensemble~\cite{dietterich00experimental,opitz99empirical,geurts06extremely,hansen90neuralensembles}. Essentially, an ensemble needs to have classifiers that are good at estimating the response in areas of the attribute space that do not overlap too much. Broadly speaking, diversity can be achieved in an ensemble by either employing different classification algorithms to train each base classifier, forming a heterogeneous ensemble; or by changing the training data or training scheme for each of a set of the same base classifier to form a homogeneous ensemble. The latter has attracted the majority of classifier ensemble research. Most often, homogeneous ensemble algorithms involve some degree of Bagging (bootstrap sampling of the training data), Boosting (iteratively re-weighting the importance of cases in the training data) and/or meta-classification such as Stacking (one classifier learns based on the outputs of classifiers lower down the stack). Popular ensemble algorithms available in the Weka toolkit\footnote{Weka:~\url{http://www.cs.waikato.ac.nz/ml/weka/}} include: Bagging decision trees~\cite{breiman96bagging}; Random Committee, a technique that creates diversity through randomising the base classifiers, which are a form of random tree; Dagging~\cite{ting97dagging};
AdaBoost (Adaptive Boosting)~\cite{freund96experiments}, which iteratively re-weights based on the training accuracy of the base classifier, usually a decision tree;
Multiboost~\cite{webb00multiboosting}, a combination of a boosting strategy (similar to AdaBoost) and Wagging, a Poisson weighted form of  Bagging; LogitBoost~\cite{friedman98logitboost}, a form of additive logistic regression; Decorate~\cite{melville04decorate}, which ensembles decision trees over real and artificially created data; Ensembles of Nested Dichotomies (END)~\cite{frank04end}, which decomposes a multiclass problem into many 2-class problems and ensembles;
 Random Forest~\cite{breiman01randomforest}, which combines bootstrap sampling with random attribute selection to construct a collection of unpruned trees; and  Rotation Forest~\cite{rodriguez06rotf}, which involves partitioning the attribute space then transforming in to the  principal components space. Of these, we think it fair to say Random Forest is by
far the most popular, and previous studies have claimed it to be amongst the most accurate of all classifiers~\cite{delgado14hundreds}.

\subsection{Heterogeneous Ensembles}

Homogeneous ensembling methods enjoy a rich literature that has produced strong classification algorithms. In contrast, advancements on heterogeneous ensembling is often the by-product of work with different main objectives, most often different methods of dividing, pruning, or combining the outputs of some given set of base classifiers, which could equally be heterogeneous or homogeneous. To an extent this is quite understandable. Generating an initial pool of heterogeneous classifiers can often be really quite arbitrary, based on either the implemented algorithms available or those that happen to be known by the researchers in question. There have however been a small number of papers directly describing schemes for forming heterogeneous ensembles. Last century,~\cite{Kittler98combining} looked at combination strategies for image data.~\cite{bagnall07meta} formulated heterogeneous ensembles for a data mining competition. An application to image classification is described in \cite{nanni15heterogeneous}, which includes an evaluation on 11 UCI data. 

These papers suggest that our central hypothesis that combining heterogeneous classifiers is worthwhile, but the sparsity of references, many of which are relatively old, indicates that the benefits are not commonly understood. Our goal is to comprehensively experimentally test this hypothesis using modern classifiers and dataset collections with a simple, transparent heterogeneous ensemble scheme in a easily reproducible way.

\subsection{Combining Classifiers}

There are many different methods for weighting and combining the outputs of a given set of ensembles members, heterogeneous or otherwise. These range from the simplest form of basic arithmetic operations~\cite{Kittler98combining} to meta-classification (stacking)~\cite{wolpert92stacking} and complex genetic and evolutionary algorithms~\cite{haque16weights}. Further, the initial base classifier set can be statically altered dataset by dataset in response to performance and/or diversity, or dynamically altered~\cite{britto14dynamic} instance to instance to generate locally optimal sub-ensembles within the problem space.

We believe that such complex schemes are not necessary to improve performance. We restrict our attention to the problem of how to combine the estimated probabilities of several classifiers after the components have been trained. This has the benefit of clarity and speed: all ensembling can be performed independently of the classifiers which can be trained concurrently. More formally, given a set of $k$ classifiers $\bm{M}=\{M_1,\ldots,M_k\}$ which produce probability estimates for any unseen case $\hat{\bm{p}}_k(\bm{x})$, the problem is to produce a final ensemble estimate $\hat{\bm{p}}$ based on weights associated with each classifier. Weighting could be of individual classifications ($\hat{y}$)  probability distributions, or probability estimates for each class. We consider weighting probabilities the simplest way of capturing the information in the output of the base classifiers. The following definitions omit the normalisation stage for clarity. Prediction weighting takes just the prediction from each member classifier,
$$\hat{p}(y=i|M,\bm{x}) \propto \sum_{j=1}^k w_j f(\hat{y}_j,i), $$
whereas probability weighting weights the distribution each classifier produces,
$$\hat{p}(y=i|M,\bm{x}) \propto \sum_{j=1}^k w_j p_j(y=i|M,\bm{x}).$$
It is common with homogeneous ensembles such as random forest to give equal weighting to all members and to combine the final predictions instead of classifiers as a whole. The approach is reasonable when there are a large number of relatively similar components since it mitigates the need for cross validation, and the only requirement for correct prediction is that on average more members predict correctly than not - a reasonable assumption given a large enough sample space of sufficiently diverse yet better-than-guessing classifiers. However, with many fewer classifiers producing very different models, simple majority vote will discard a large amount of useful information.

\section{HESCA: the Heterogeneous Ensembles of Standard Classification Algorithms}
\label{hesca}
HESCA is intentionally as simple as we could make it. It sums each classifier's exponentially weighted probability distributions. Training (Algorithm~\ref{algo:build}) consists of finding a weight for each classifier based on cross validation of the train data, before building each classifier on the full train data. We effectively treat each classifier as a black box. If internal model selection or parameter tuning is needed as part of any classifier's training, it occurs independently on each cross validation fold in {\bf findWeight} and also again on the full train data in {\bf buildClassifier}.

\begin{algorithm}[!ht]
 	\caption{HESCA Train Classifier(A train set $D_r$)}
	\label{algo:build}
	\begin{algorithmic}[1]
 \REQUIRE A set of classifiers $\{M_1,\ldots, M_k\}$
\ENSURE	A set of trained classifiers $\{M_1,\ldots, M_k\}$ and weights $\{w_1,\ldots, w_k\}$
\FOR {$i\leftarrow 1$ to $k$}
		\STATE $w_i\leftarrow M_i.\text{findWeight}(D_r)$ \COMMENT{Cross validate for weight}
        \STATE $M_i.\text{buildClassifier}(D_r)$
	\ENDFOR
 	\end{algorithmic}
 \end{algorithm}
Classification involves forming a combined probability distribution (Algorithm~\ref{algo:classify}).
\begin{algorithm}[!ht]
 	\caption{HESCA Distribution for Instance (A test case $\bm{x}$)}
	\label{algo:classify}
	\begin{algorithmic}[1]
 \REQUIRE A set of classifiers $<M_1,\ldots, M_k>$, an exponent $\alpha$, a set
 of weights, $w_i$ and the number of classes $c$
 \ENSURE Probability estimates for each class, $\hat{\bm{p}}$	
	\STATE $\hat{\bm{p}}={0,\ldots,0}$ \COMMENT{final c probabilities for classifier} 	
	\FOR {$i\leftarrow 1$ to $k$}
		\STATE $\hat{\bm{q}} \leftarrow M_i.\text{distributionForInstance}(\bm{x})$
		\FOR {$j\leftarrow 1$ to $c$}
			\STATE $\hat{p}_j \leftarrow \hat{p}_j+ w_i^\alpha \cdot \hat{q}_{i}$
		\ENDFOR
	\ENDFOR
	\STATE $s \leftarrow 0$ \COMMENT{normalise}
	\FOR {$i\leftarrow 1$ to $c$}
		\STATE$s\leftarrow s+ \hat{p}_i$
	\ENDFOR
	\FOR {$i\leftarrow 1$ to $c$}
		\STATE$\hat{p}_i\leftarrow \hat{p}_i/c$
	\ENDFOR
 	\end{algorithmic}
 \end{algorithm}
We have intentionally not tried to optimise the classifiers within HESCA, since our whole thesis is that it is easy to leverage off the diversity of different algorithms that are about the same on average. We have made two design decisions with HESCA: the choice of weighting mechanism (accuracy) and the decision to exponentiate the weight $\alpha$, which we use to attenuate differences in accuracy.

The weight could be a function of any of the performance metrics described in Section~\ref{stats} (error, balanced error, log likelihood or AUROC), or alternatives such as precision, recall, their combination the F-Score, Confusion Entropy~\cite{wei10evaluating} and Mathews Correlation Coefficient~\cite{matthews75lyozme}. We have experimentally compared these measures (with $\alpha$ set to 1 for all) and accuracy was not significantly worse than any of the rest. Based on our guiding principle of simplicity, we chose to weight by accuracy.

As $\alpha$ increases, the weightings of classifiers found to be stronger on the training data relative to the rest are increased, until the ensemble becomes functionally identical to the single best classifier in training. Conversely, when alpha is 0 all members will be equally weighted. To simplify further, by removing the need to tune $\alpha$ and potentially overfitting, we fix $\alpha$ to 4 for all experiments and all component structures. We chose this exact value fairly arbitrarily as a sensible starting point.

\begin{figure}[htb]
    \label{fig:alpha}
    \centering
          {\includegraphics[width =\linewidth, trim={1.5cm 1.5cm 1.5cm 1.5cm},clip]{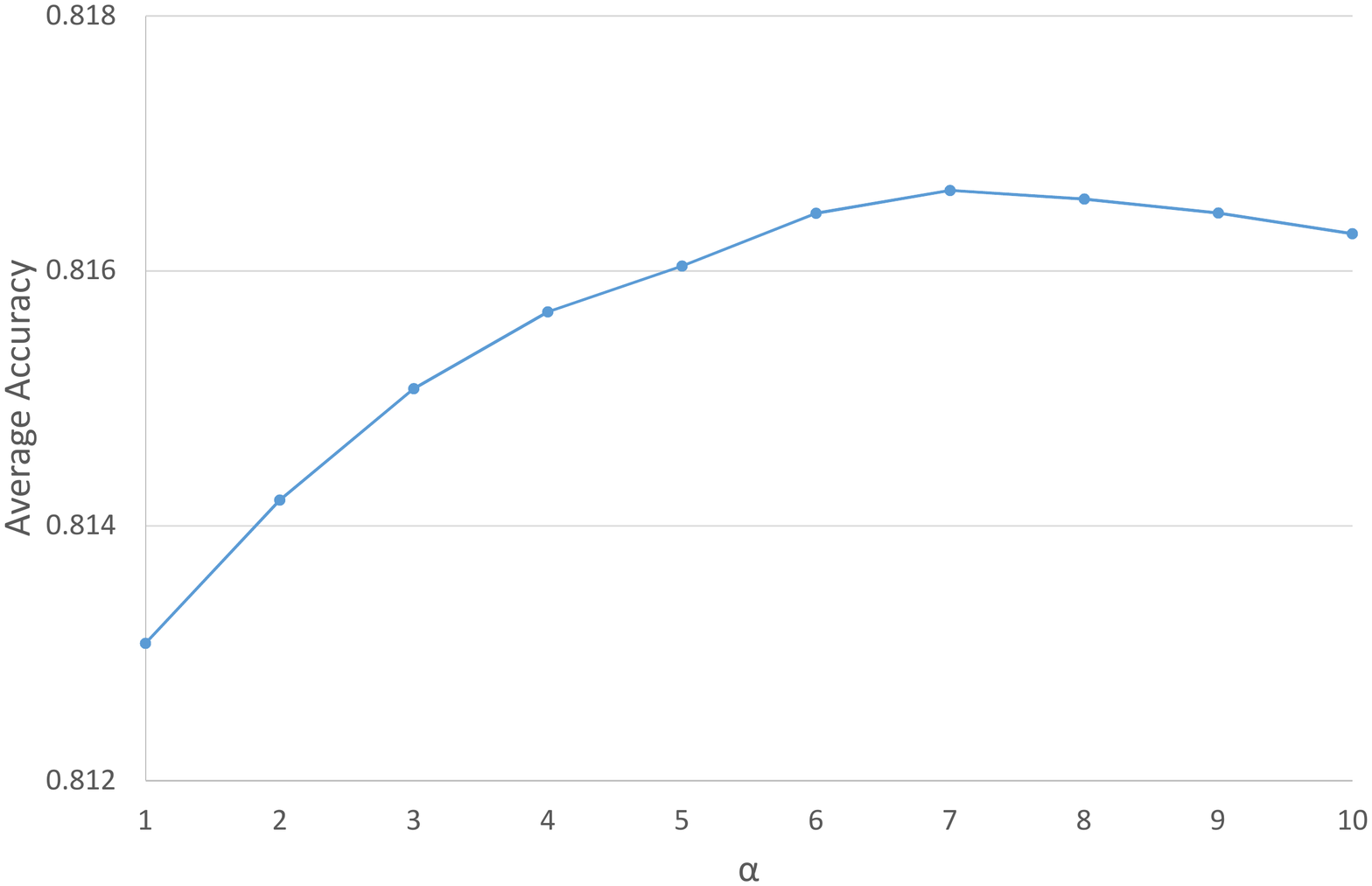}}
    \caption{The average test accuracy over 121 UCI data sets (each data set sampled 30 times) of HESCA with weighting parameter $\alpha$ between 1 and 10. The components are the basic classifiers described in Section~\ref{betterClassifier}}
\end{figure}

Later experiments indicate that there may be some consistent benefit in setting alpha higher or by cross validation. Figure~\ref{fig:alpha} shows the average accuracy over UCI data sets of a HESCA classifier for $\alpha$ values from 1 to 10. Accuracy seems to peak around $\alpha=7$. However, the differences are very small, and while a similar trend is found on the UEA-UCR datasets, these were generated for only a single set of components. To avoid any risk of overfitting we continued with $\alpha=4$ for all experiments.

The key hypothesis we wish to test is whether, given a set of classifiers that are approximately as accurate as each other on average, does using HESCA improve performance in relation to the components? We look at two variants of HESCA. The first, called just HESCA, contains the following five classifiers: logistic regression (Logistic); C4.5 decision tree (C4.5); linear support vector machine (SVML); nearest neighbour classifier (NN); and a multi layer perceptron (MLP), with a single hidden layer. These were chosen because they are well known, commonly used, relatively fast to train, conceptually diverse, and we believed {\em a priori} there would be little difference between them. This last factor lead us to exclude naive Bayes, which in our experience tends to perform poorly on problems with just real valued attributes. There are stable implementations of these five classifiers in the Weka toolkit, which allows us to provide a simple Weka HESCA classifier. The Weka version of HESCA can be used as a standalone classifier (building all the components internally) or it can combine the outputs of other classifiers.

The second version, HESCA+, contains four classifiers commonly considered to be state-of-the-art. These are a Random Forest (RandF), a Rotation Forest (RotF), a support vector machine with Quadratic kernel (SVMQ), and a deep neural network (with two hidden layers) (DNN). All classifiers in HESCA+ are implemented in Weka, with the exception of the DNN. There is currently no option in Weka to use an MLP with more than one hidden layer so we have used Keras\footnote{Keras:~\url{https://keras.io/}} and TensorFlow\footnote{TensorFlow:~\url{https://www.tensorflow.org/}} for the DNN. Our goal is not to assess DNN for classification; we wish to do the minimum to create a decent classifier not significantly worse than the other HESCA+ components. However, training a DNN with default parameters is highly unlikely to achieve this goal. Initialising and optimising hyperparameters for deep models is of critical importance to their performance. We tune the DNN based on recommendations from the literature.

We optimise 3 parameters: the learning rate (from 0.1 to 0.00001 on a log$_{10}$ scale), the number of nodes in the first hidden layer (from the range of $1.5m$ to $5m$, where $m$ is the number of attributes), and the number of nodes in the second layer (from the number of class values to the number of nodes in the first hidden layer). As per the recommendations in~\cite{bengio12practical} we use stochastic gradient descent with momentum (with momentum fixed to 0.9~\cite{krizhevsky12imagenet}) and we do not use a learning rate schedule as~\cite{bengio12practical} states {\em ``in many cases the benefit of choosing other than this default value is small"}. We use a random grid search~\cite{bergstra12random} when training, giving each model 20 parameter options, and each is evaluated using a 3-fold cross validation on the training data only with early stopping criteria when the model processes 100 epochs without an increase in hold-out accuracy. The best parameter setting from the training experiment is then applied to the final model, using all training data to build and the same number of epochs derived from the training cross validation.


\section{Results on UCI Data}
\label{results}
We have conducted hundreds of million experiments to test the central hypothesis related to HESCA that on average, HESCA makes its components better. Here we present condensed results concisely and without further analysis or breakdown to avoid obfuscating our key contributions. In Section~\ref{analysis} we break down these results and investigate why HESCA makes components better.

Experiments are conducted on averages over 30 stratified resamples of data, with 50\% of the data taken for training, 50\% for testing. All classifiers are aligned on the same folds. These are reproducible using the method \texttt{(InstanceTools.resampleInstances(dataset,foldNumber,0.5)}, or alternatively all folds can be downloaded\footnote{~\url{http://research.cmp.uea.ac.uk/HESCA/UCIContinuous.zip} and ~\url{http://research.cmp.uea.ac.uk/HESCA/UCIContinuousFolds.zip} (3.5 GB)}. HESCA is implemented in Java using Weka. DNN is implemented in TensorFlow. All code is available and open source\footnote{~\url{http://research.cmp.uea.ac.uk/HESCA/large17hescaCode.zip}}. The experiments can be reproduced (see class \newline
\texttt{vector\_classifiers.HESCA}). In the course of experiments we have generated gigabytes of prediction information and results. These are available in raw format and in summary spreadsheets~\footnote{~\url{http://research.cmp.uea.ac.uk/HESCA/large17hescaResults.zip} and ~\url{hescaAllResults.zip} (9 GB)}.

Section~\ref{betterClassifier} demonstrates that both versions of HESCA are significantly better than their components. Whilst gratifying, our natural skepticism makes us wonder if we have not just discovered a result that could easily be reproduced in another way. We consider the following possible explanations: Can we get equivalent results by simply choosing a classifier rather than ensembling (Section~\ref{chooseBest})? Can we get equivalent results by tuning a single classifier rather than using HESCA (Section~\ref{tuning})? Why not just use a homogeneous ensemble (Section~\ref{homos})? And is the result just an artifact of the components of the versions of HESCA we use (Section~\ref{components})?

\subsection{Does HESCA improve equivalent base classifiers?}
\label{betterClassifier}

\begin{figure}[!ht]
	\centering
\begin{tabular}{cc}
       \includegraphics[width =6cm, trim={2cm 7cm 1cm 5cm},clip]{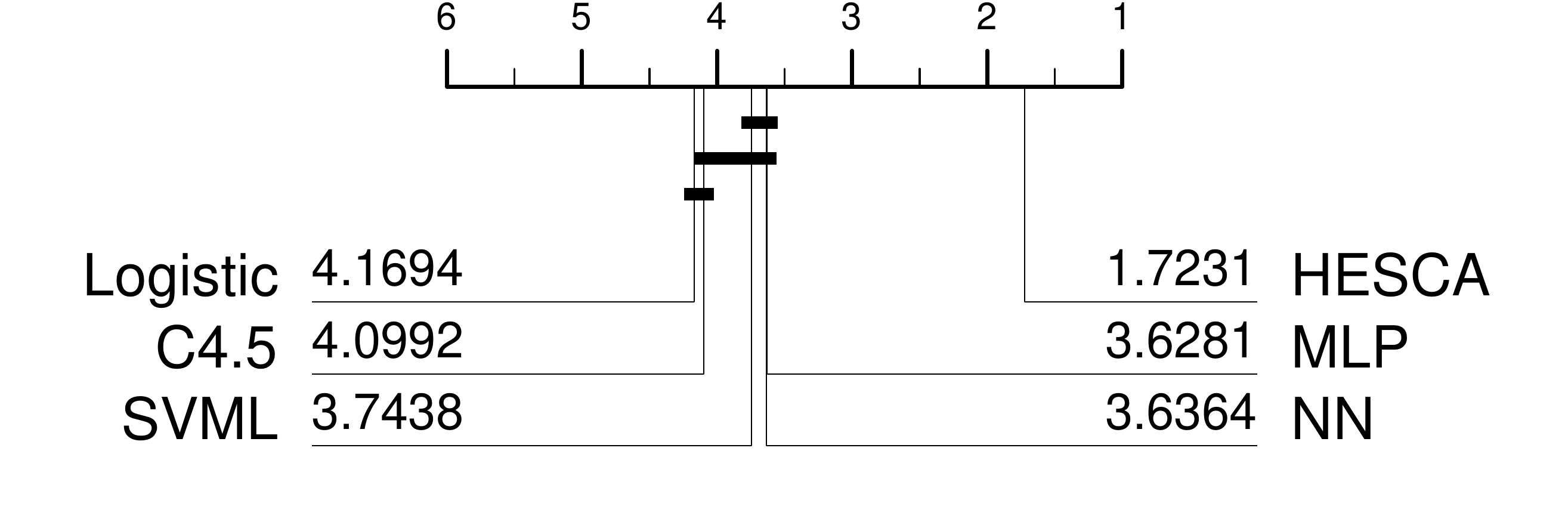}              	
&
       \includegraphics[width =6cm, trim={2cm 7cm 1cm 5cm},clip]{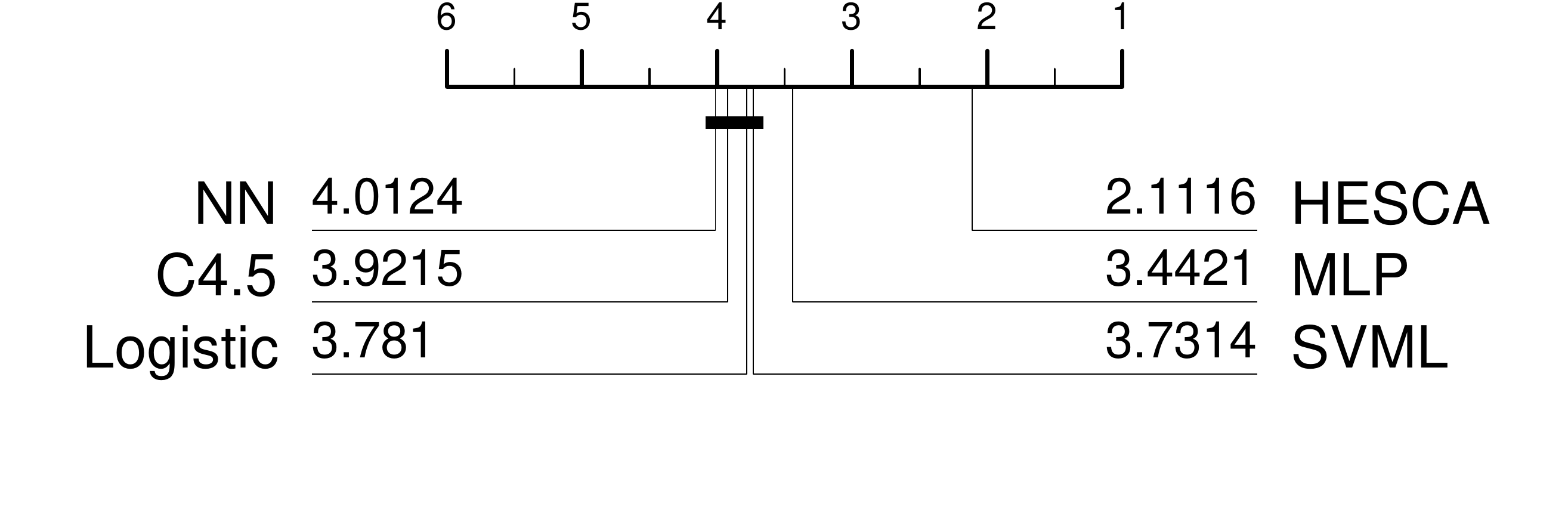}  \\
(a) Error & (b) Balanced Error \\
       \includegraphics[width =6cm, trim={2cm 7cm 1cm 4cm},clip]{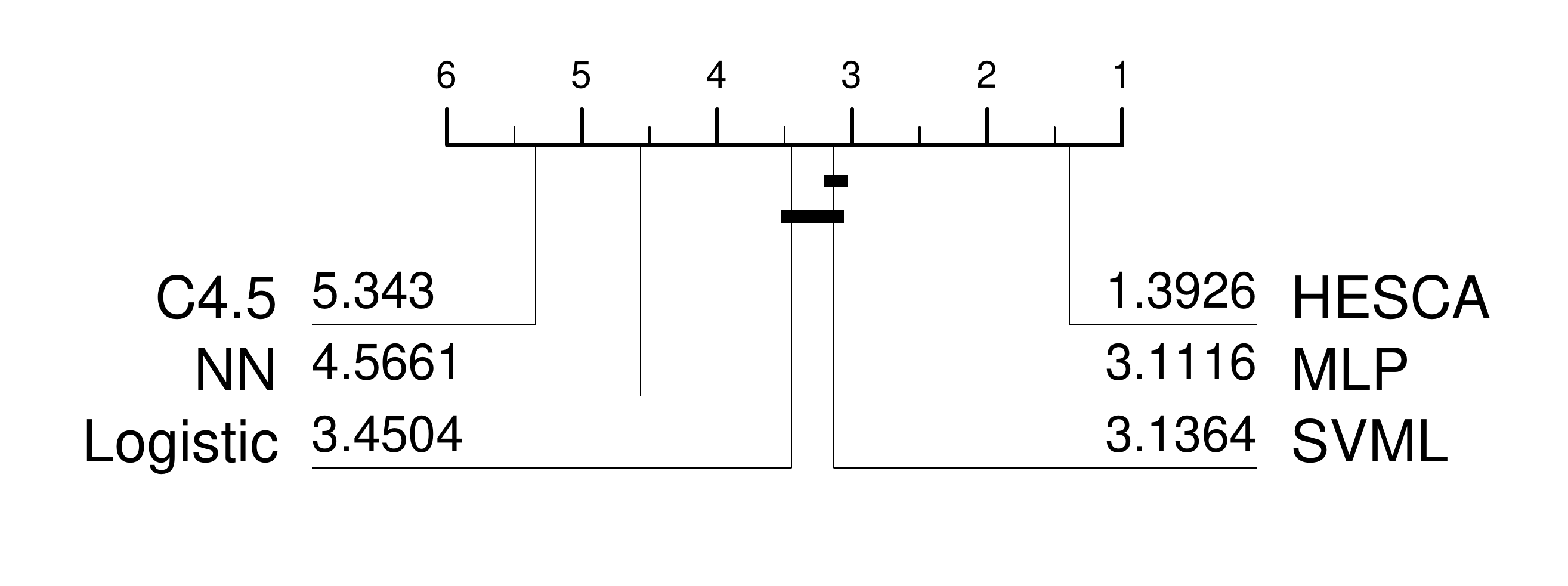}
&
       \includegraphics[width =6cm, trim={2cm 7cm 1cm 4cm},clip]{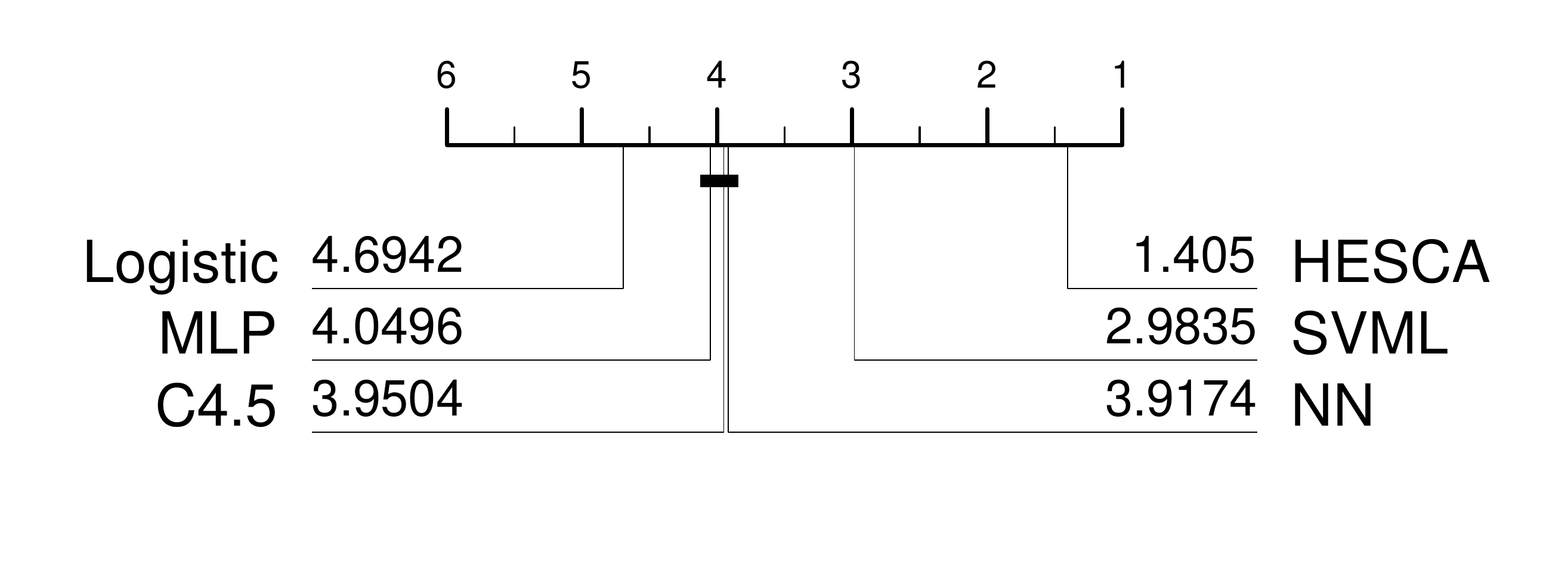}              	\\
(c) AUROC & (b) NLL \\          
       \end{tabular}
       \caption{Critical difference diagrams for HESCA with basic classifiers on the UCI data.}
       \label{hescaUCI}
\end{figure}

Figure~\ref{hescaUCI} shows the critical difference diagrams for HESCA on the 121 UCI datasets. Figures~\ref{hescaUCI}(a) and ~\ref{hescaUCI}(b) show there is very little difference between the five basic classifiers in terms of either error measure, but that HESCA has significantly lower error. This is solid evidence to support our base hypothesis. Figure~\ref{hescaUCI}(c) shows HESCA is significantly better at relative ordering of the test data, as measured by AUROC. In terms of the components, it is curious that C4.5 and NN have significantly worse AUROC than the other three components, but the NLL is not significantly different. We can think of no obvious reason for this. Figure~\ref{hescaUCI}(d) shows HESCA produces significantly better probability distribution estimates than its members. We note the surprising fact that logistic regression is significantly worse than SVML, which uses logistic regression to form probability distributions from the support vectors. It is beyond the scope of this work to tease out reasons for minor differences in classifier performance. However, the variation between Figures~\ref{hescaUCI}(a), (b), (c) and (d) does reinforce the value of using alternative metrics. The fact is that HESCA is significantly better on average for all four statistics. When we compare performance over folds for each problem, we once again see the benefit of HESCA. If we perform a paired two sample t-test on each data set, we find that HESCA has significantly lower error than the best performing component (MLP) on 86 of the 121 data sets, and significantly higher error on just 3 datasets.

\begin{figure}[!ht]
	\centering
\begin{tabular}{cc}
       \includegraphics[width =6cm, trim={3.5cm 2cm 1.5cm 2.5cm},clip]{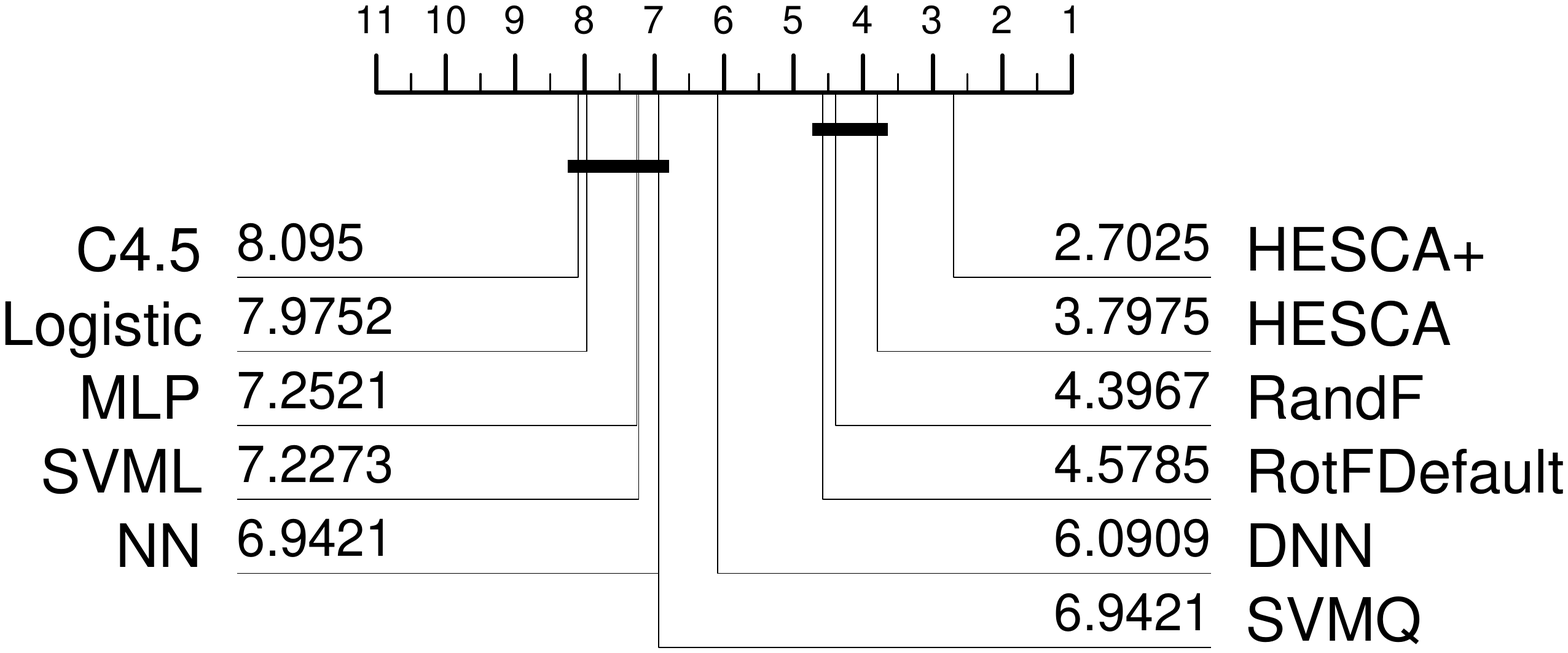}              	
&
       \includegraphics[width =6cm, trim={3.5cm 2cm 1.5cm 2.5cm},clip]{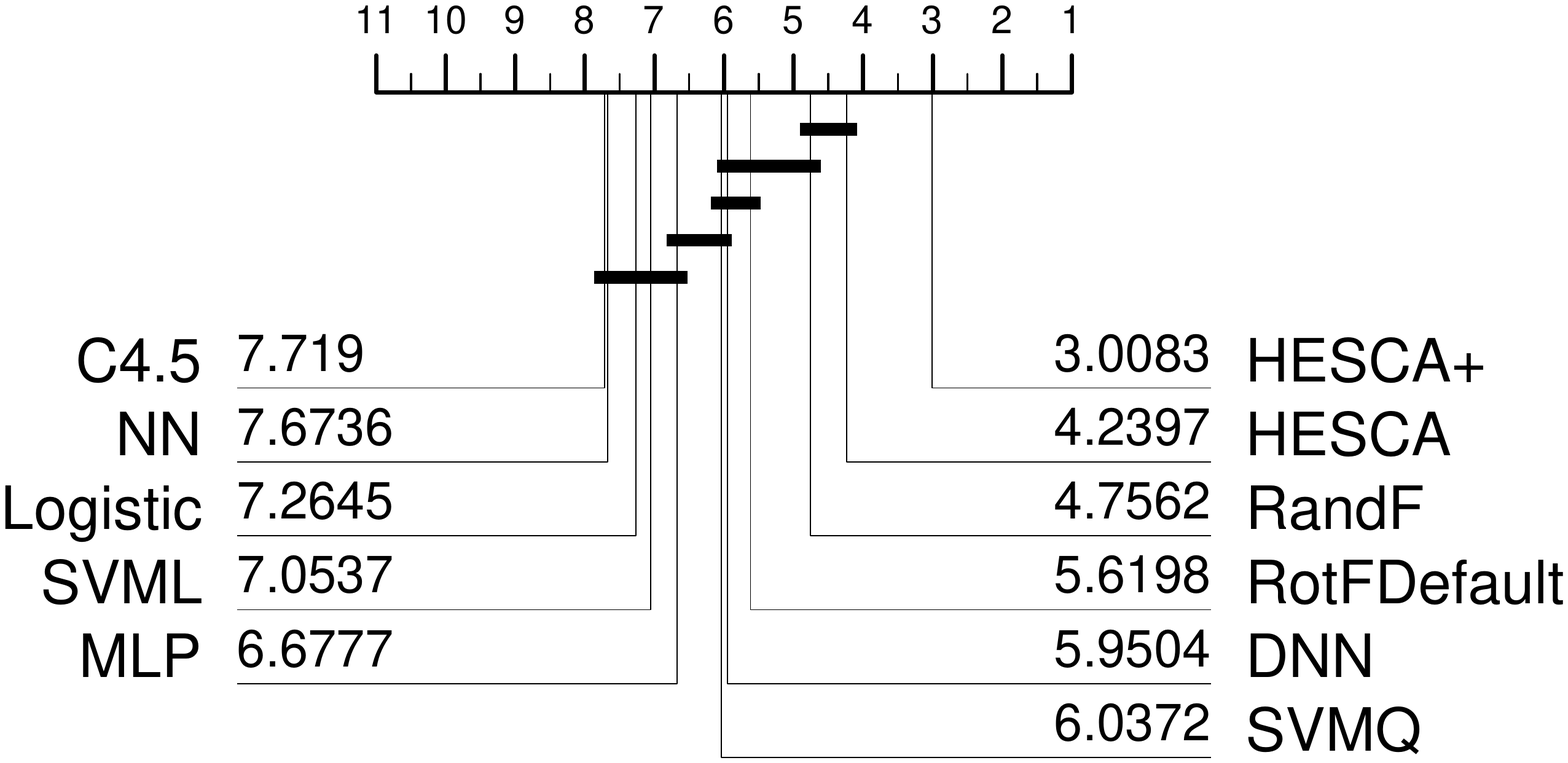}  \\
(a) Error & (b) Balanced Error \\
       \includegraphics[width =6cm,trim={3.5cm 2cm 1.5cm 2.5cm},clip]{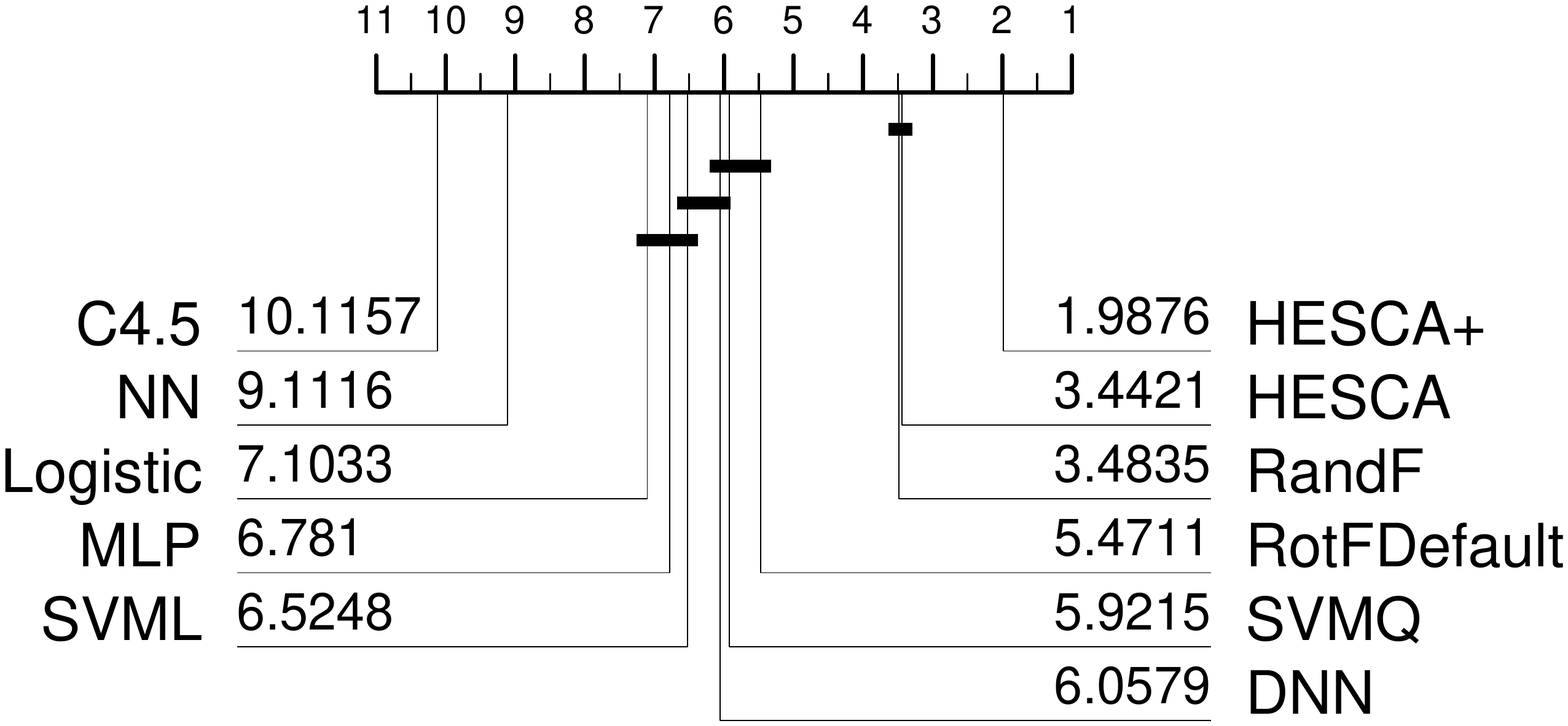}
&
       \includegraphics[width =6cm,trim={3.5cm 2cm 1.5cm 2.5cm},clip]{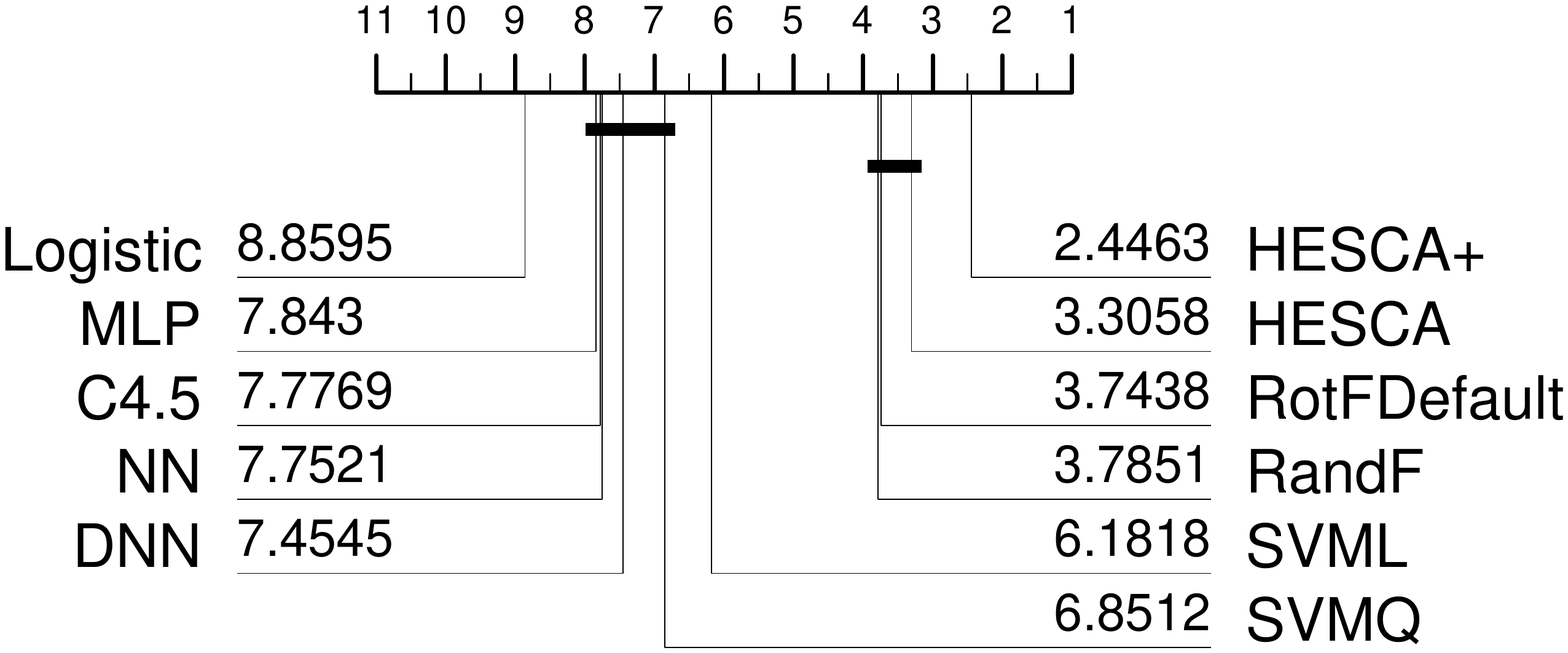}              	\\
(c) AUROC & (d) NLL \\
       \end{tabular}
       \caption{Critical difference diagrams for HESCA+ on the UCI data.}
       \label{hesca+UCI}
\end{figure}

It could be argued that making the basic classifiers in HESCA better is not of great interest, since more sophisticated algorithms will probably be better. We could counter that it is not always possible to build an advanced classifier, but generally would concede the point. The experiments described in Figure~\ref{hescaUCI} were conceived largely as a test of concept and the quality of HESCA as a classifier surprised us. Nevertheless, on most problems, the practitioner has enough computing power to run a range of more modern algorithms such as support vector machines, random forest or deep neural networks. HESCA+ contains examples of these three families of algorithm (described in Section~\ref{hesca}). Figure~\ref{hesca+UCI} shows the critical difference diagrams for the five base classifiers in HESCA, the four components of HESCA+ and the two HESCA variants. The primary conclusion from these diagrams is that on average HESCA+ is significantly better than its components. We note that Random Forest is the best performing algorithm, which agrees with previous experimental results~\cite{delgado14hundreds} and that the forest algorithms are significantly better than SVMQ and DNN. However, we stress that our goal is not to test which is the best component and acknowledge that we could have probably made the components better through parameter tuning. We address the issue of improving components through tuning in Section~\ref{tuning}. It is of interest, however, that HESCA is not significantly different to random forest on any of the four metrics we consider.


The crucial observation is that both configurations of HESCA give significant improvement over their components. We would argue that, based on these experiments and other published results, HESCA is as good a classifier  as the current state-of-the-art and HESCA+ represents an advance in classification algorithms or real valued attributes. We now investigate whether we could achieve the same improvement through an alternative experimental scheme.

\subsection{Is it better to just choose a classifier using the error estimates from the train data?}
\label{chooseBest}

\begin{figure}[!ht]
\begin{tabular}{cc}
       \includegraphics[width =6cm, trim={2cm 13.5cm 1cm 2.5cm},clip]{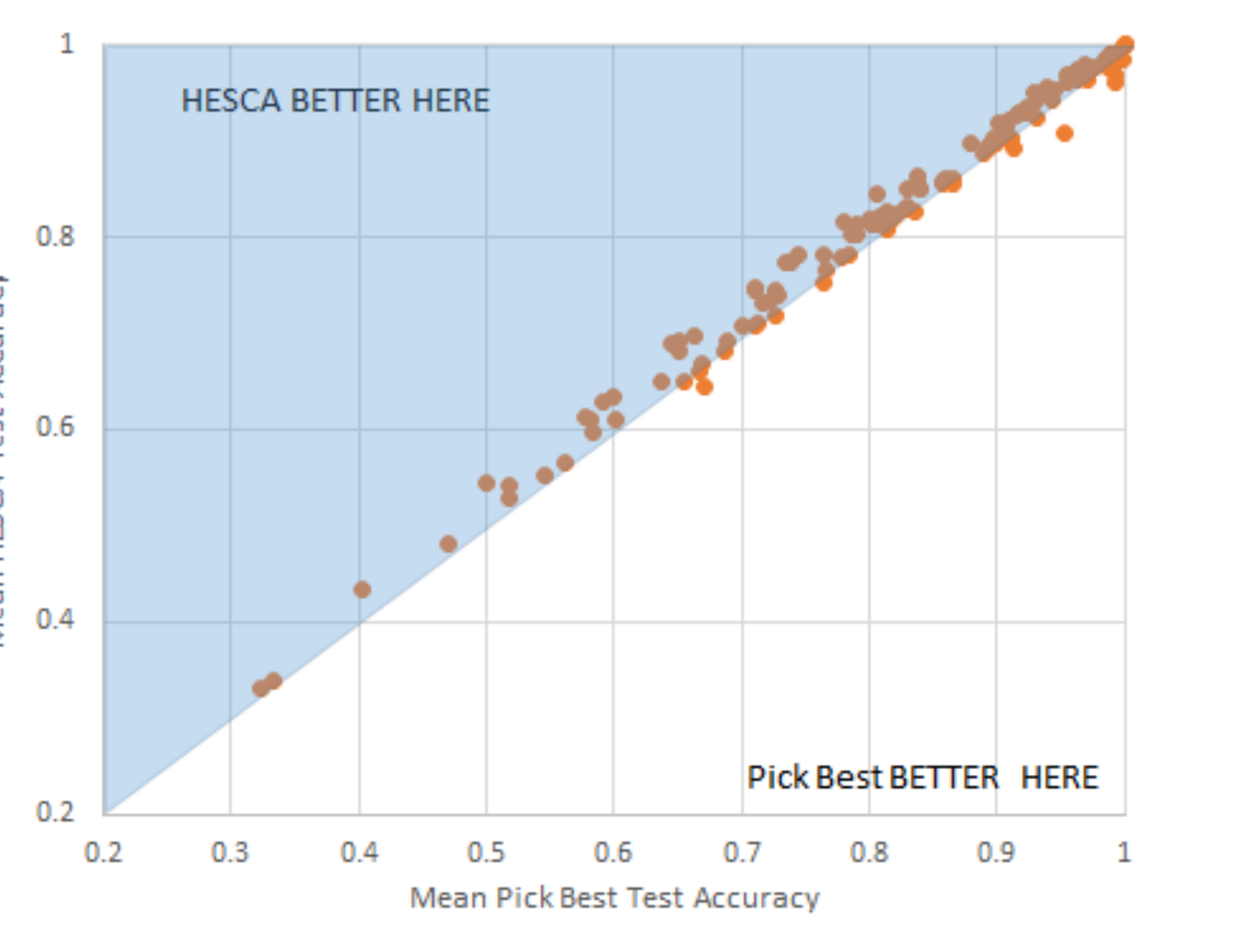}
&
       \includegraphics[width =6cm, trim={2cm 13.5cm 1cm 2.5cm},clip]{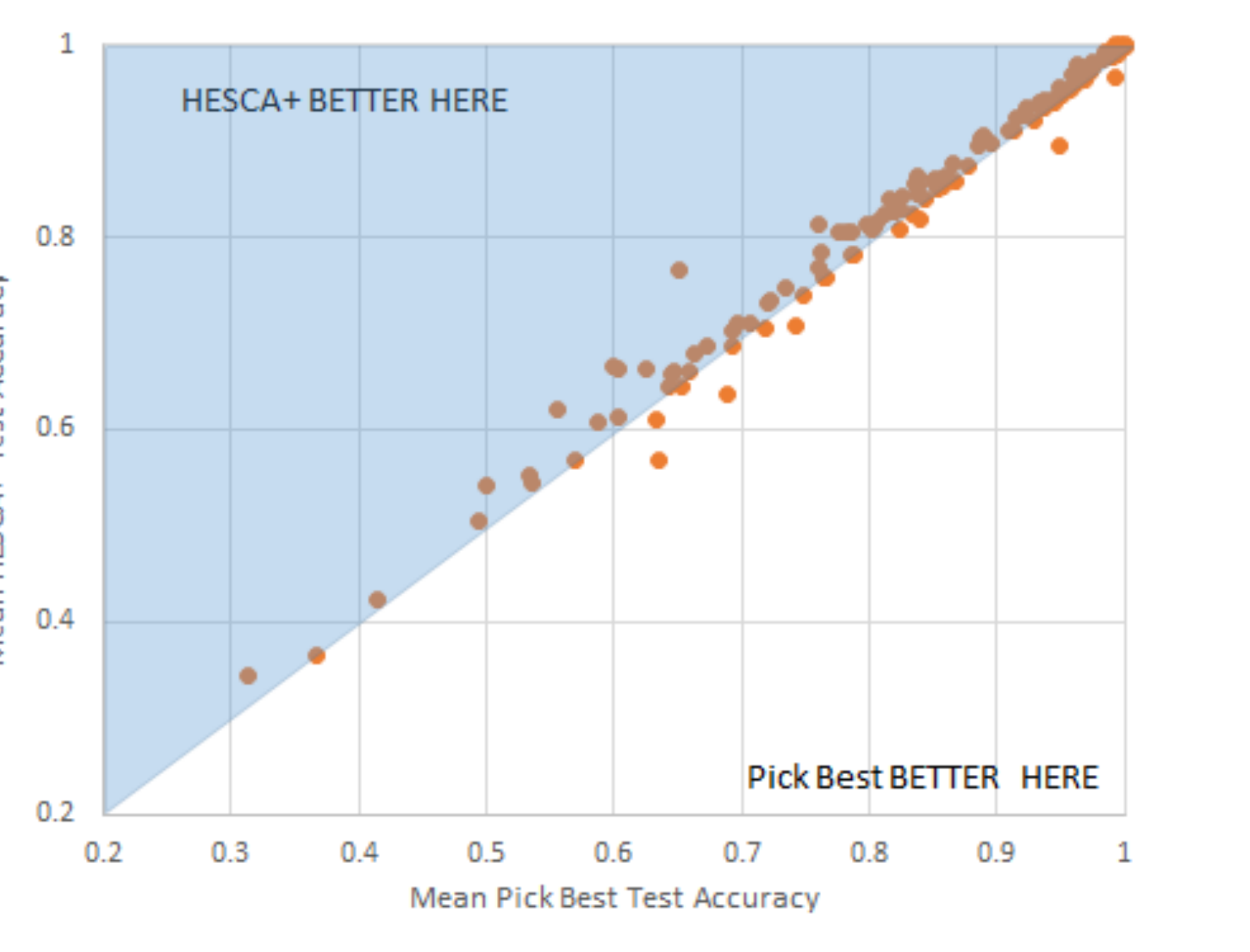} \\
       (a) & (b)
       \end{tabular}
       \caption{(a) Accuracy of HESCA vs pick best component and (b) HESCA+ vs pick best component.}
       \label{pickBestUCI}
\end{figure}

Given HESCA ensembles based on estimates of accuracy obtained from the train data, it seems reasonable to ask, why not just choose the classifier with the highest estimate of accuracy? The answer is that, because of the variance in the accuracy estimate, it is on average significantly worse choosing a single classifier than using the HESCA ensembles. Figure~\ref{pickBestUCI} shows the scatter plots of accuracy for choosing the best base classifier from their respective component sets against using HESCA and HESCA+. On average over 30 folds, HESCA is better on 81 data, pick best on 37 and they tie on 3. HESCA+ is better on 78, pick best on 40 and they tie on 3. The differences are significant.

\begin{figure}[!ht]
\begin{center}
       \includegraphics[width =10cm, trim={2cm 2cm 2cm 2cm},clip]{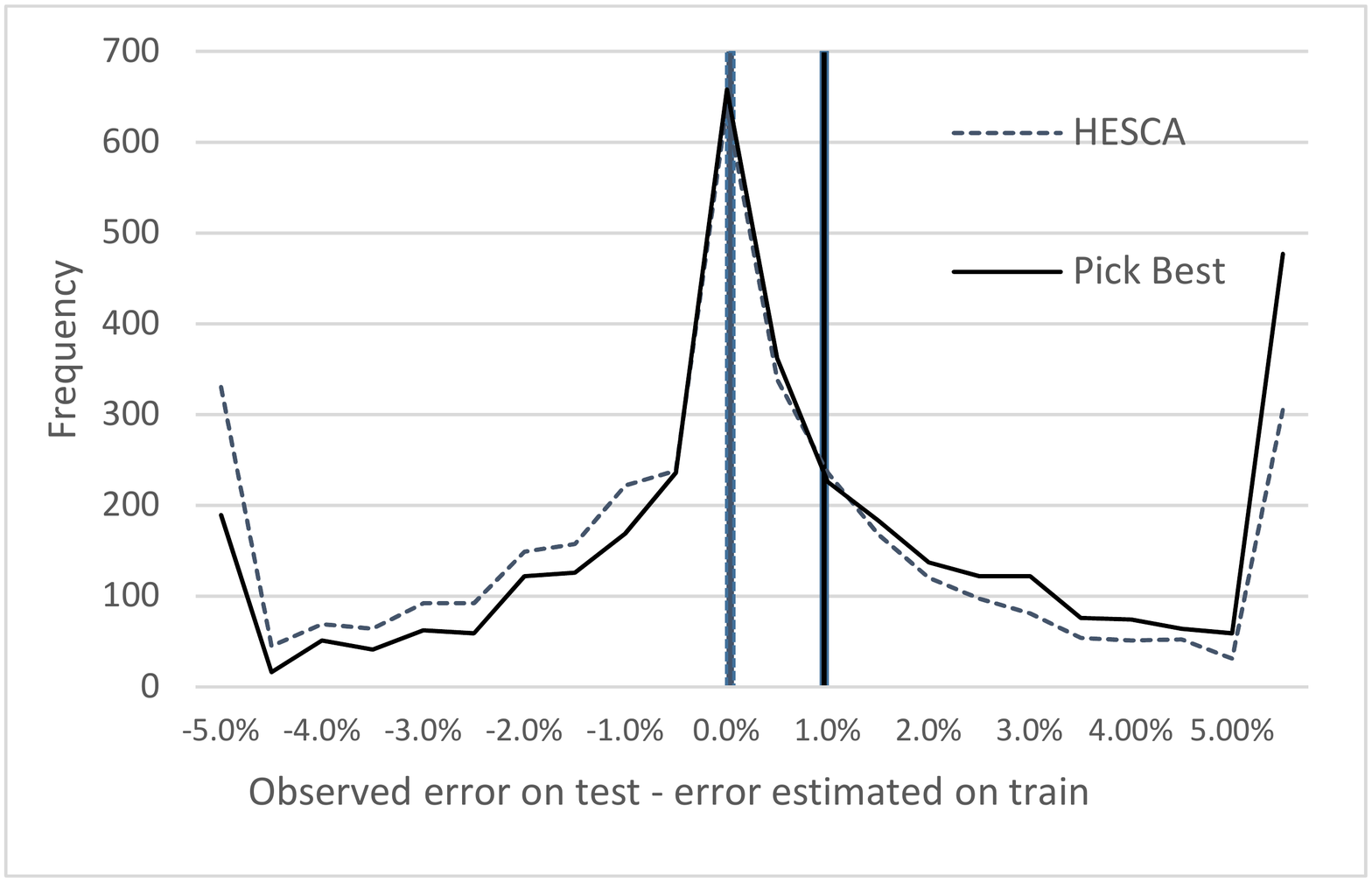}
       \caption{Distribution of observed bias over 3630 folds of the UCI data. Solid lines represent the means over all observations. Pick best underestimates the error rate by 1.12 on average; HESCA over-estimates it by 0.18.}
       \label{trainTestUCI}
\end{center}
\end{figure}

We explore whether this can be explained by the characteristics of the data in Section~\ref{analysis}. Another reason for ensembling rather than choosing the best is that you get a much better estimate of the test error from the train data with HESCA without the need for a further level of cross validation. Suppose we compare the difference in the estimated error from train data and the observed test error. A consistent difference would indicate bias, with a positive difference meaning train error is consistently underestimated. Figure~\ref{trainTestUCI} shows the distribution of the bias taken over all 3630 folds of the UCI data. Pick Best tends to underestimate the error; HESCA tends to overestimate it. However, overall, HESCA bias is on average insignificant, whereas Pick Best underestimates error by 1.12\%.

When comparing algorithms over entire archives, we get a good sense of those which are better for general purpose classification. However, it could be the case that HESCA is just more consistent that its components: a jack of all trades ensemble that achieves a high ranking most of the time, but is usually beaten by one or more of its components. A more interesting improvement is an ensemble that consistently achieves higher accuracy than all of its components. For this to happen, the act of ensembling needs to not only cover for the weaknesses of the specialists in suboptimal domains, but accentuate their strengths within their specialisation also.

\begin{figure}[!ht]
	\centering
      \includegraphics[width =10cm, trim={1cm 1cm 1cm 1cm},clip]{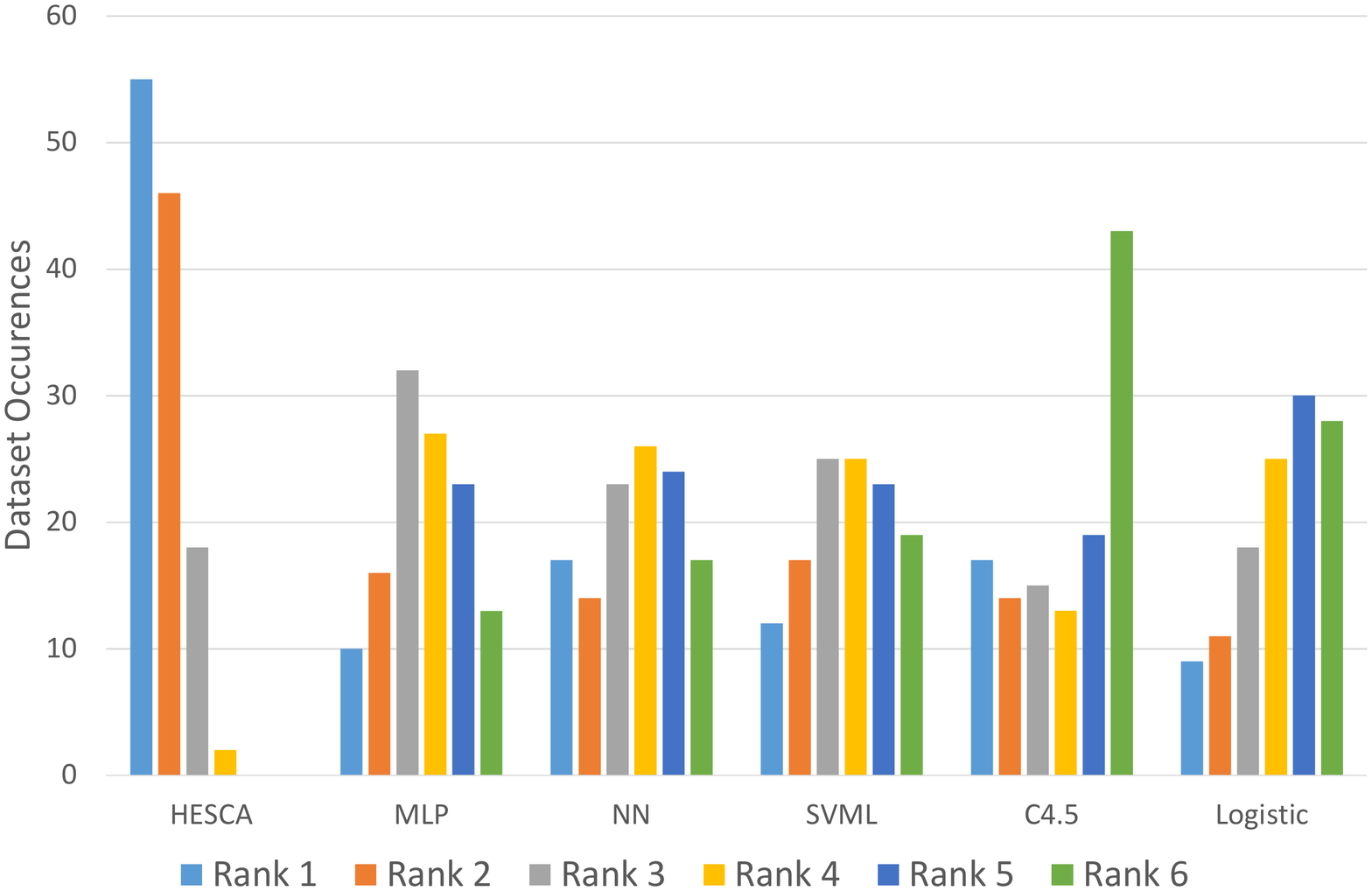}
      \caption{Histograms of accuracy rankings over the 121 UCI datasets for HESCA and its components.}
      \label{rankingsHESCA}
\end{figure}
\begin{figure}[!ht]
	\centering
      \includegraphics[width =10cm, trim={1cm 1cm 1cm 1cm},clip]{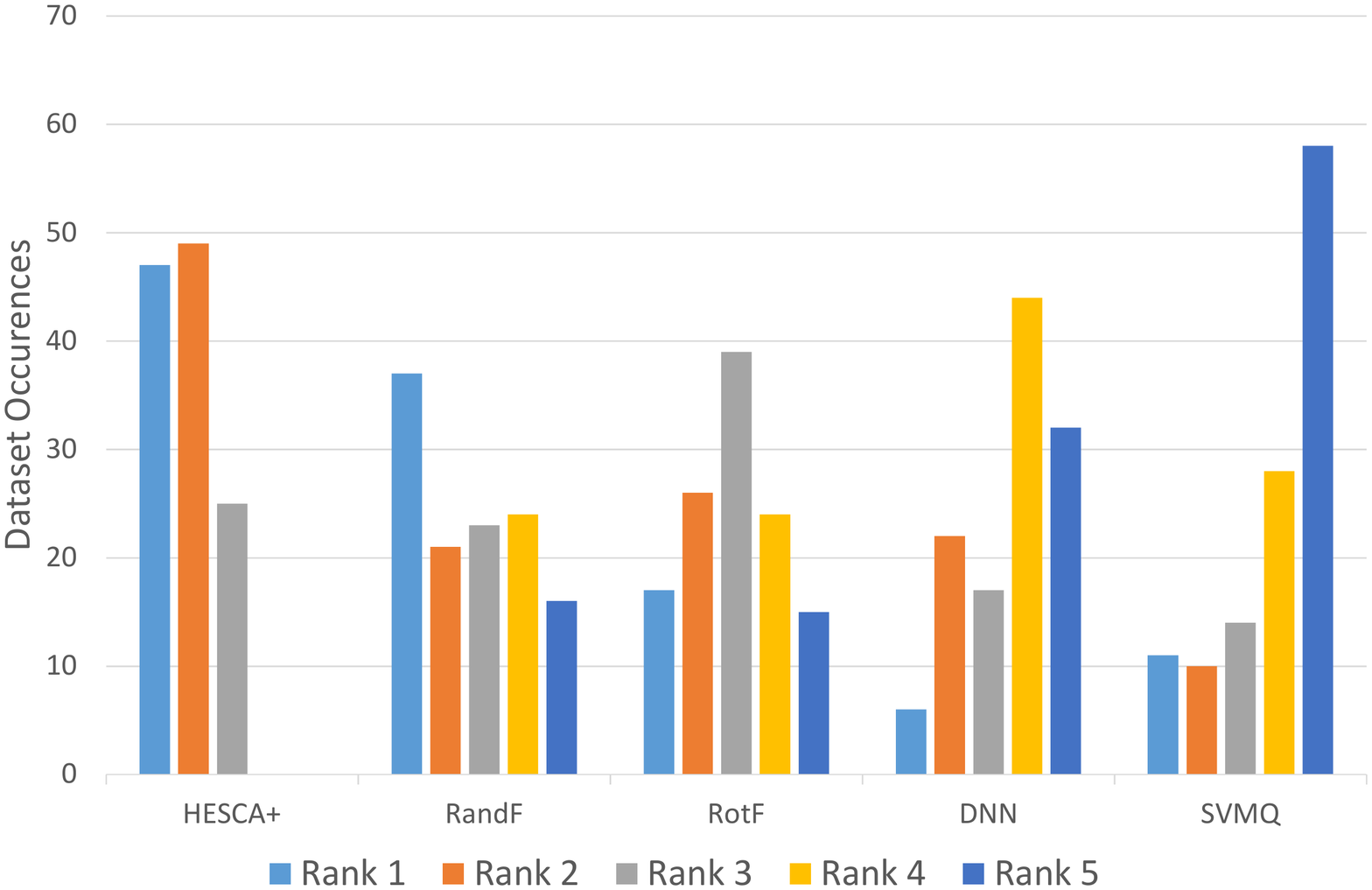}
      \caption{Histograms of accuracy rankings over the 121 UCI datasets for HESCA and its components.}
      \label{rankingsHESCA+}
\end{figure}
Figure~\ref{rankingsHESCA} shows the counts of the rankings achieved by HESCA and its components, in terms of accuracy, over the 121 UCI datasets. HESCA is the single best classifier far more often than any of its components, and is in fact more often the best classifier than second best. HESCA also is never ranked fifth or sixth, and is ranked fourth only twice, demonstrating the consistency of the improvement. This suggests that the simple combination scheme used in HESCA is able to actively enhance the predictions of its locally specialised members, rather than just achieve a consistently good rank. Figure~\ref{rankingsHESCA+} shows the same data for HESCA+ and components. HESCA+ is ranked first or second on the vast majority of datasets, and is never ranked fourth or fifth.

\subsection{Is it better to tune a single classifier rather than use HESCA?}
\label{tuning}
With the exception of DNN, where some tuning is essential, both HESCA and HESCA+ use untuned classifiers.  However, tuning parameters on the train data can significantly improve classifier accuracy~\cite{bagnall17muppets}. This begs the obvious question: would a carefully tuned classifier do as well or better than HESCA and HESCA+? To investigate whether this is the case, we tune a SVM (known to be particularly sensitive to tuning) using the spherical Radial Basis Function (TunedSVMRBF). We perform a ten-fold cross validation for the parameters $(C,\gamma) \in \{(2^{-16},2^{-16}), (2^{-16},2^{-15}), \ldots, (2^{16},2^{16}) \}$. Ten-fold cross validation on 1089 different parameter combinations over 30 folds gives a total number of 326,700 models for every data set. For the slowest data set (miniboone), sequential execution would take more than 6 months. However, we can distribute folds and parameter combinations over a reasonably sized cluster. Even so, considerable computation is required, and we were unable to complete a full parameter search for 4 datasets (within a 7 day limit): adult; chess-kvrk; miniboone; and magic. To avoid bias, we perform this analysis without these results. On average, both HESCA and HESCA+ are significantly better than TunedSVMRBF in terms of error, balanced error, NLL and AUROC. The mean difference in average error between TunedSVMRBF and HESCA/HESCA+ is 0.5\% and 1.5\% respectively. HESCA has lower error than TunedSVMRBF on 61\% of problems, HESCA+ on 68\%. We investigate these results further in Section~\ref{analysis}. However, we believe that, by taking a classifier widely considered one of the best and tuning it over a very large parameter space, we have shown that the positive results for HESCA cannot be explained by the lack of tuning of the components. Even with orders of magnitude more computational train time, TunedSVMRBF is significantly worse than both HESCA and HESCA+. It could be the case that an alternative SVM configuration and parameter search technique does better, but our discussions with experts in SVM suggest our approach is not unreasonable. Even if we could configure a SVM to do as well as HESCA or HESCA+, the computational time is likely to be far greater for the SVM. Sequential execution of HESCA for miniboone (including all internal cross validation) is under 8 hours, and for HESCA+ it is three days. HESCA can build all but 6 of the datasets in under an hour. On average, if we were to sequentially execute the classifiers, HESCA is two orders of magnitude faster than the tuned SVMRBF and HESCA+ is one order of magnitude faster. 
We conclude that it is not possible to dismiss the HESCA results as being an artifact of not tuning the base classifiers.

\subsection{Are any of the existing homogeneous ensembles better than HESCA?}
\label{homos}

In Section~\ref{enembles} we identified 11 alternative homogeneous ensembles. Given we have already seen that two of them,  random forest and rotation forest, are not significantly worse than HESCA (see Figure~\ref{hesca+UCI}), it seems fair to evaluate the other 9 homogeneous ensembles. We ran these classifiers on the UCI datasets using the Weka default values. We acknowledge the danger of using default parameters~\cite{bagnall17muppets}, but there is a limit to the number of experiments we can reasonably perform and believe homogeneous ensembles are generally robust to the most important parameter, number of base classifiers, as long as this is fairly large.

\begin{figure}[!ht]
	\centering
\begin{tabular}{cc}
       \includegraphics[width =6cm, trim={1.5cm 4cm 0.5cm 2.5cm},clip]{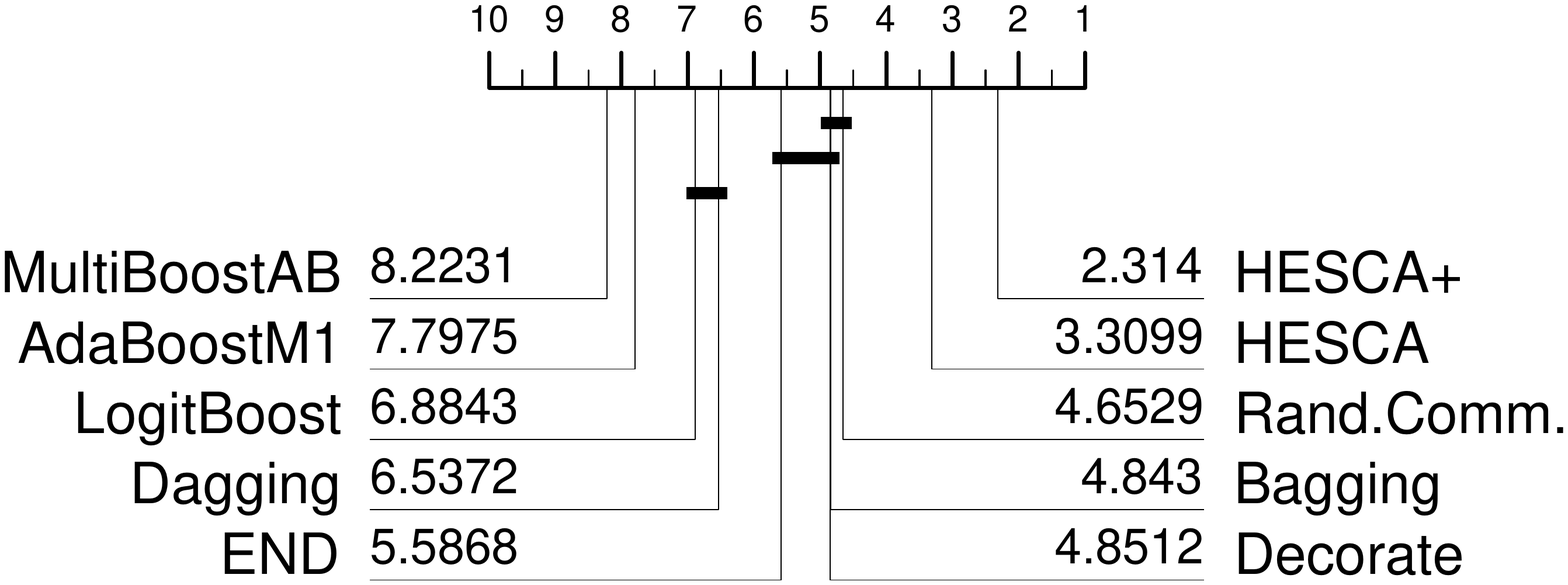}              	
&
       \includegraphics[width =6cm, trim={1.5cm 4cm 0.5cm 2.5cm},clip]{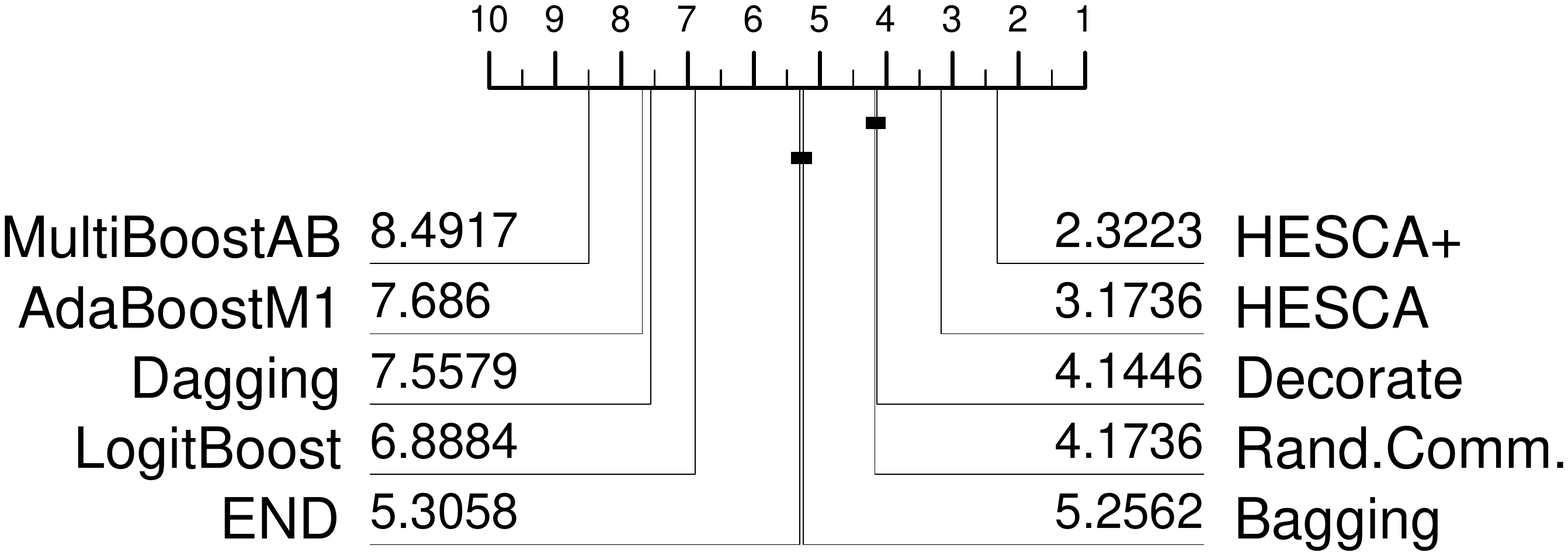}  \\
(a) Error & (b) Balanced Error \\
       \includegraphics[width =6cm,trim={1.5cm 4cm 0.5cm 2.5cm},clip]{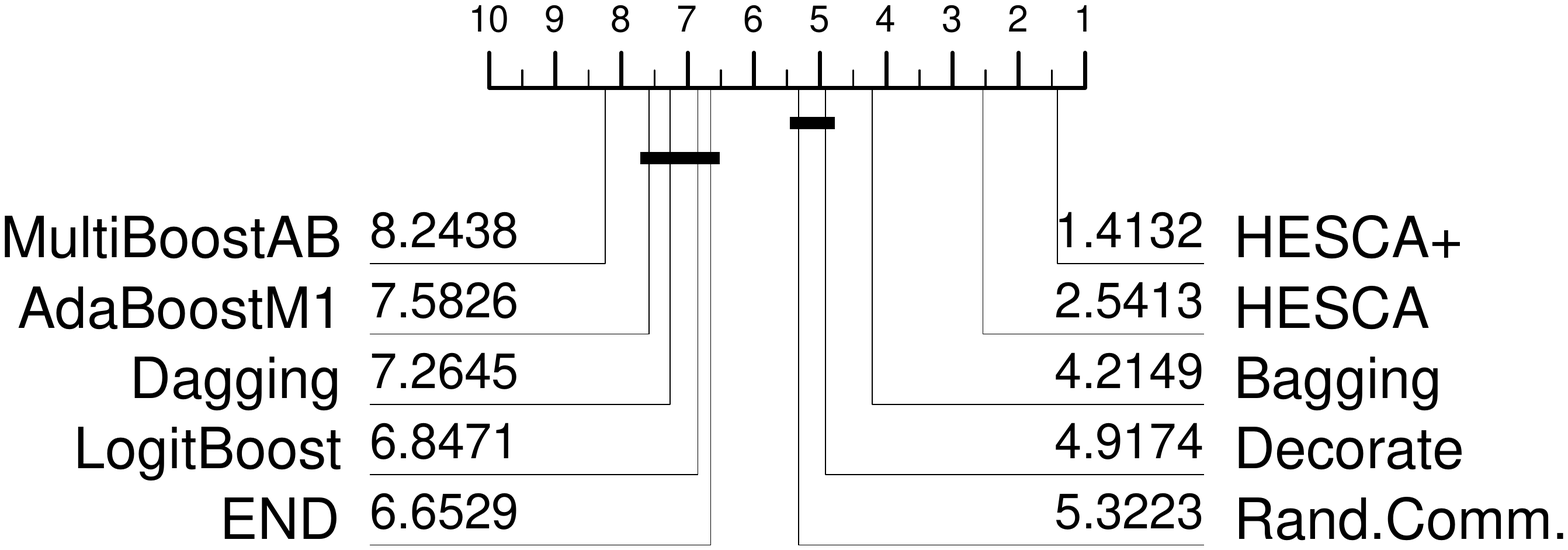}
&
       \includegraphics[width =6cm,trim={1.5cm 4cm 0.5cm 2.5cm},clip]{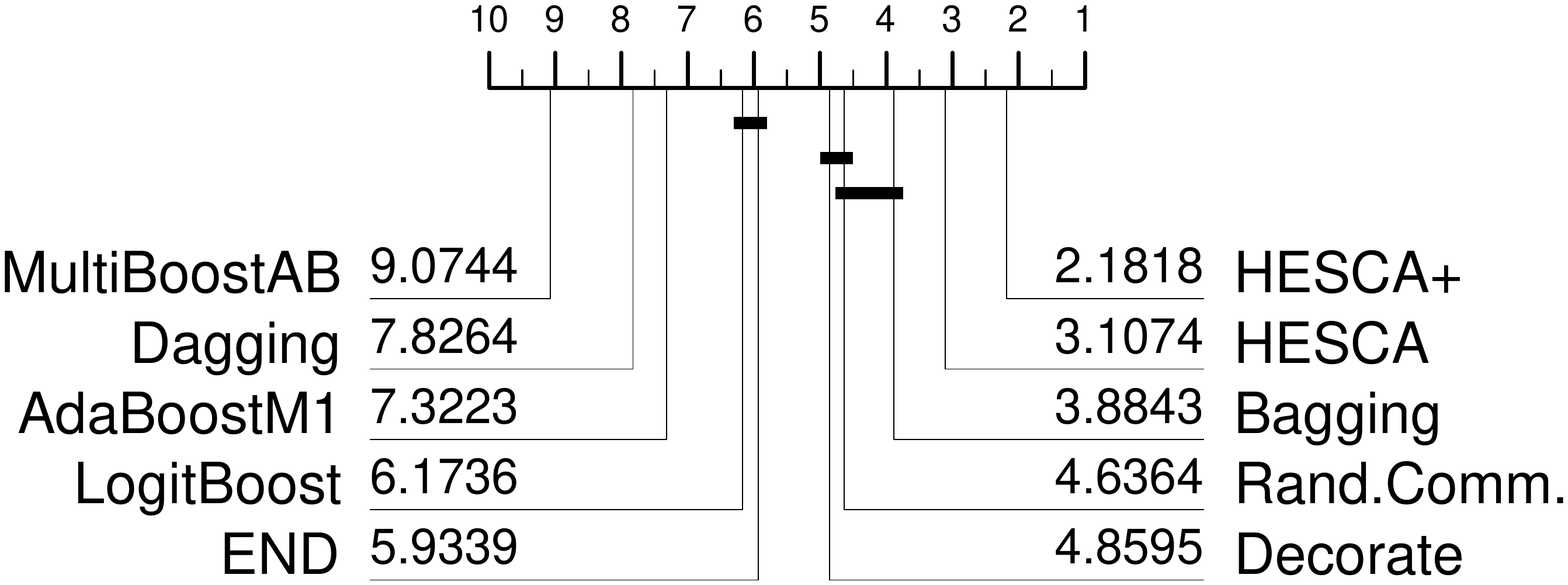}              	\\
(c) AUROC & (b) NLL \\
       \end{tabular}
       \caption{Critical difference diagrams for homogeneous ensembles and  HESCA.}
       \label{otherEnsembles}
\end{figure}

Figure~\ref{otherEnsembles} shows the results of 9 homogeneous classifiers, HESCA and HESCA+. We observe that HESCA and HESCA+ are significantly more accurate than the other ensembles. This is surprising, given the huge amount of research effort into designing homogeneous ensembles and the relatively little attention paid to heterogeneous ensembles. It suggests that the sampling of data, diversification of attributes and combining the outputs in clever ways is less important than the nature of the classifiers in the ensemble.

\subsection{Is it the particular configuration that makes HESCA better than its components?}
\label{components}

\begin{figure}[!ht]
	\centering
\begin{tabular}{cc}
       \includegraphics[width =6cm, trim={1.5cm 6.5cm 0cm 5cm},clip]{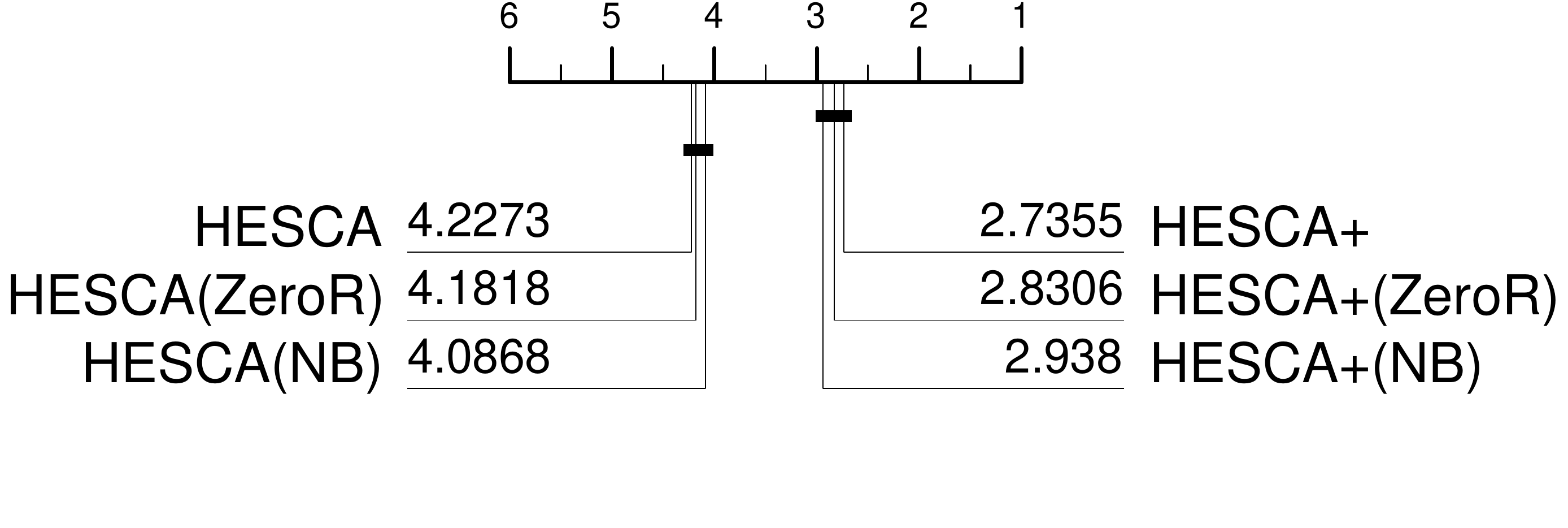}              	
&
       \includegraphics[width =6cm, trim={1.5cm 6.5cm 0cm 5cm},clip]{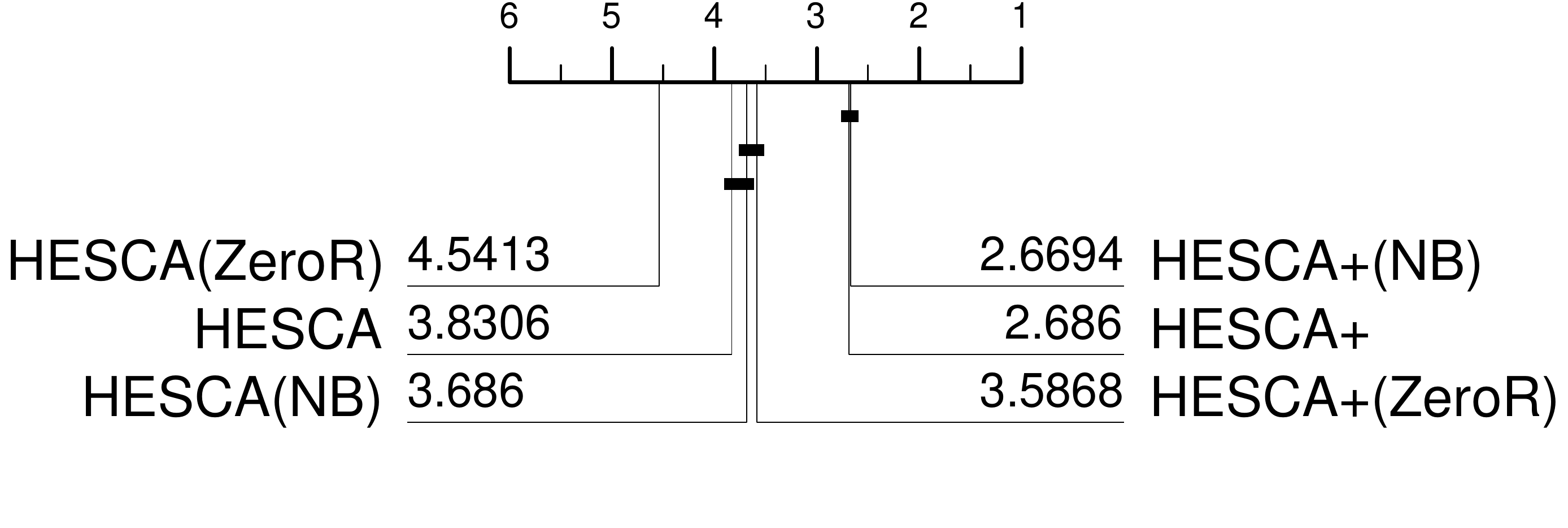}  \\
(a) Error & (b) Balanced Error \\
       \includegraphics[width =6cm,trim={1.5cm 6.5cm 0cm 4cm},clip]{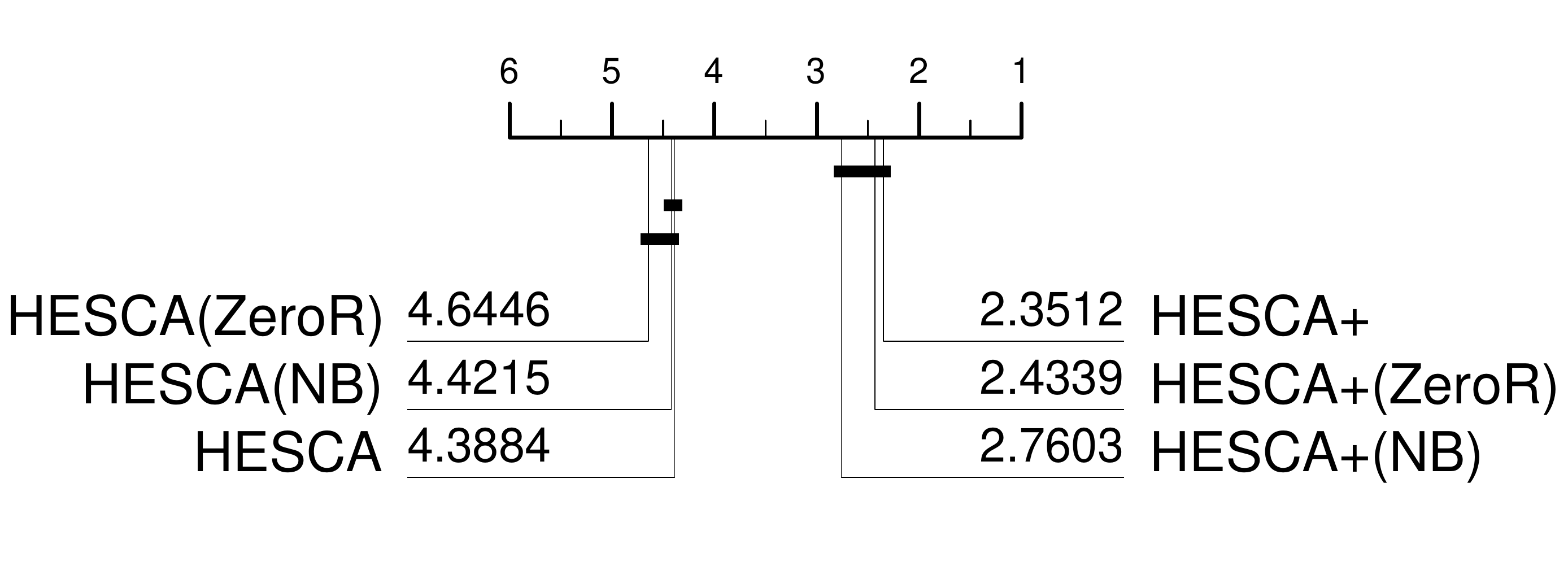}
&
       \includegraphics[width =6cm,trim={1.5cm 6.5cm 0cm 4cm},clip]{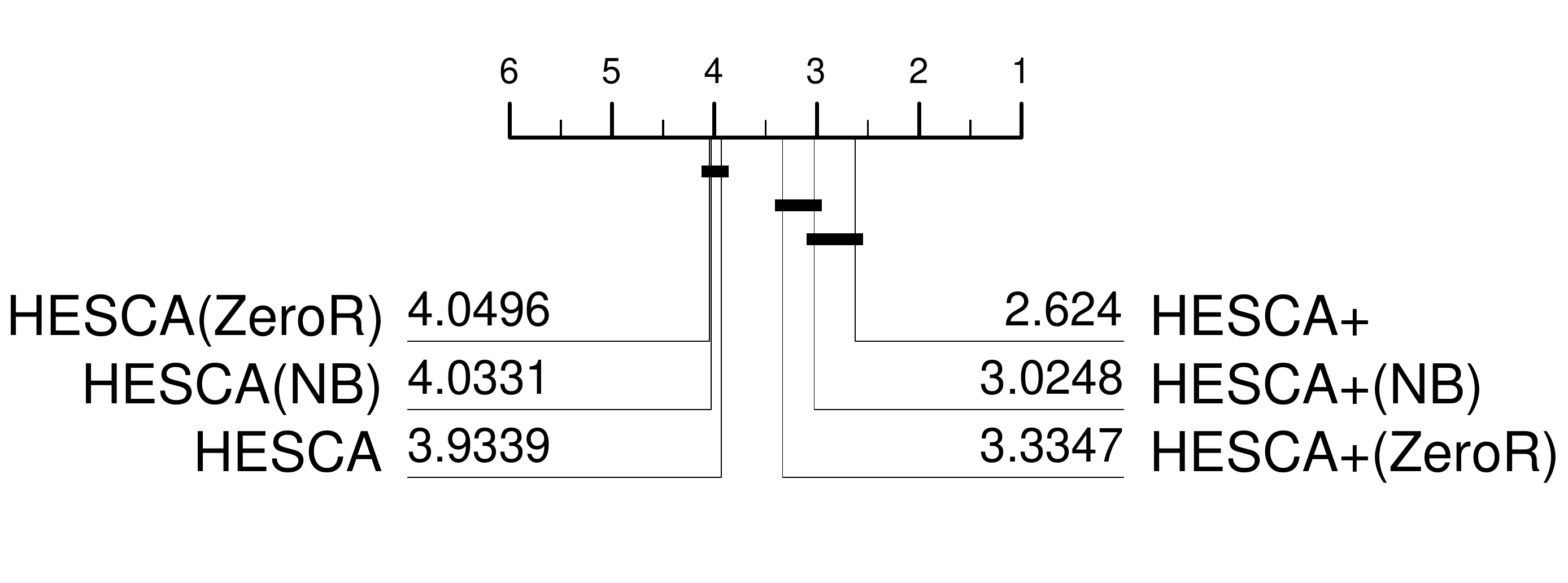}              	\\
(c) AUROC & (b) NLL \\
       \end{tabular}
       \caption{Critical difference diagrams for HESCA and HESCA+ with weak classifiers zeroR and Naive Bayes (NB) added.}
       \label{badHesca}
\end{figure}

It is worth considering how sensitive HESCA is to the component classifiers. Does adding a classifier much worse than the others make the overall HESCA worse? To test this we add the ZeroR classifier, which always predicts the majority class, and the Weka naive Bayes classifier that from experience we know to perform poorly on problems with only real valued attributes. Figure~\ref{badHesca} summarises the results. Adding zeroR does not significantly alter HESCA or HESCA+ in terms of error, which is our primary statistic of interest, or AUROC. Adding ZeroR to HESCA and HESCA+ make both significantly worse in terms of balanced error, and HESCA+ worse at estimating probabilities, which, given the nature of ZeroR, is unsurprising. Nevertheless, we consider the results in Figure~\ref{badHesca} demonstrate the robustness of the weighting scheme to the occasional bad classifier.

Another possible explanation for the significant improvement of HESCA over its components is that it is just a result of the classifiers we chose to use rather than a general principle.
In the course of these experiments, we have built over 22 different classifiers on the same resamples of the UCI data (see Table~\ref{tab:allClassifiers} for a list of algorithms for which we have a full set of results). Because HESCA can be post processed directly from stored results, we can use these files to test our base hypothesis that HESCA improves components that are not significantly different to each other.

\begin{table}
\caption{All the classifiers fully evaluated on the UCI datasets. All apart from the deep neural network are the standard Weka implementations. }
\label{tab:allClassifiers}
\begin{tabular}{lll}
 k-nearest neighbour  &  Decision table	      &  Naive Bayes       \\
 Rep tree             &  Decorate	            &  Random Forest    \\
1-nearest neighbour   &  Deep neural network	 &   RandomCommittee  \\
AdaBoostM1            &  END	                 &    Rotation Forest\\
Bagging	              &  Logistic    	        &   SVM (linear kernel)    \\
Bayesian Network      &  LogitBoost	          &   SVM (quadratic kernel)\\
C4.5 decision tree    &	MultiBoostAB	      &     \\
Dagging	              &  Multilayer Perceptron&	 \\
\end{tabular}
\end{table}

We randomly sampled 5 classifiers and constructed a HESCA variant (we denote the generic ensemble over any components as HESCA* to avoid confusion). Over 200 random configurations, HESCA* was significantly better than the best component on 143 (71.5\%). Note that many of these variants contain components that are significantly different, with average accuracies ranging all the way between 81.4\% and 62.7\%.



\begin{figure}[!ht]
	\centering
\begin{tabular}{cc}
       \includegraphics[width =6cm, trim={3cm 9cm 0cm 4cm},clip]{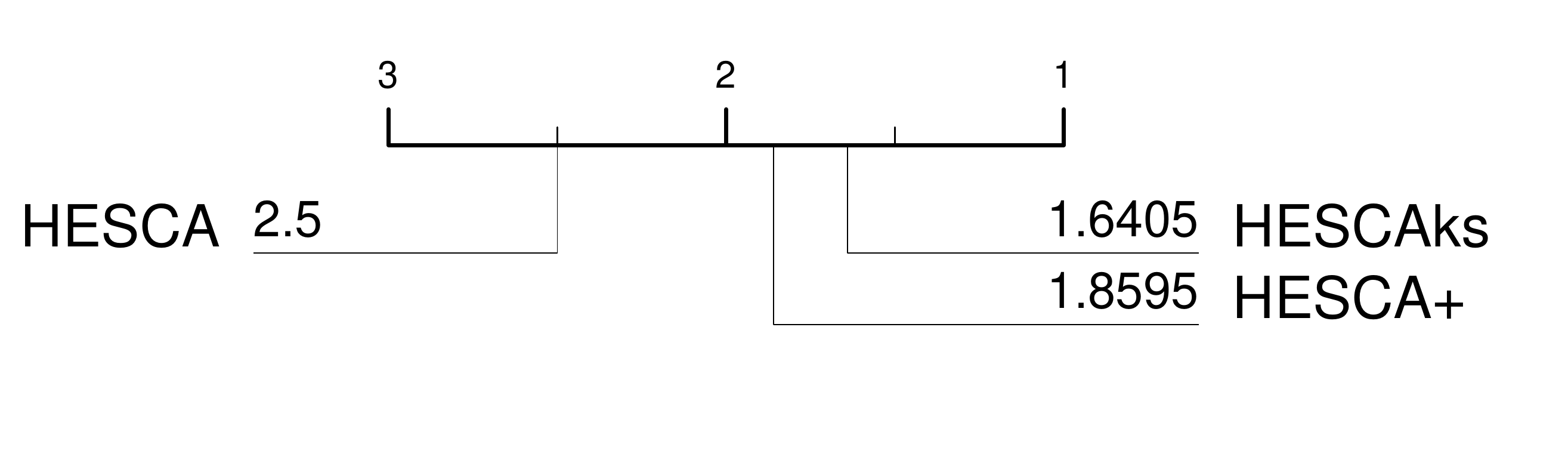}              	
&
       \includegraphics[width =6cm, trim={3cm 9cm 0cm 4cm},clip]{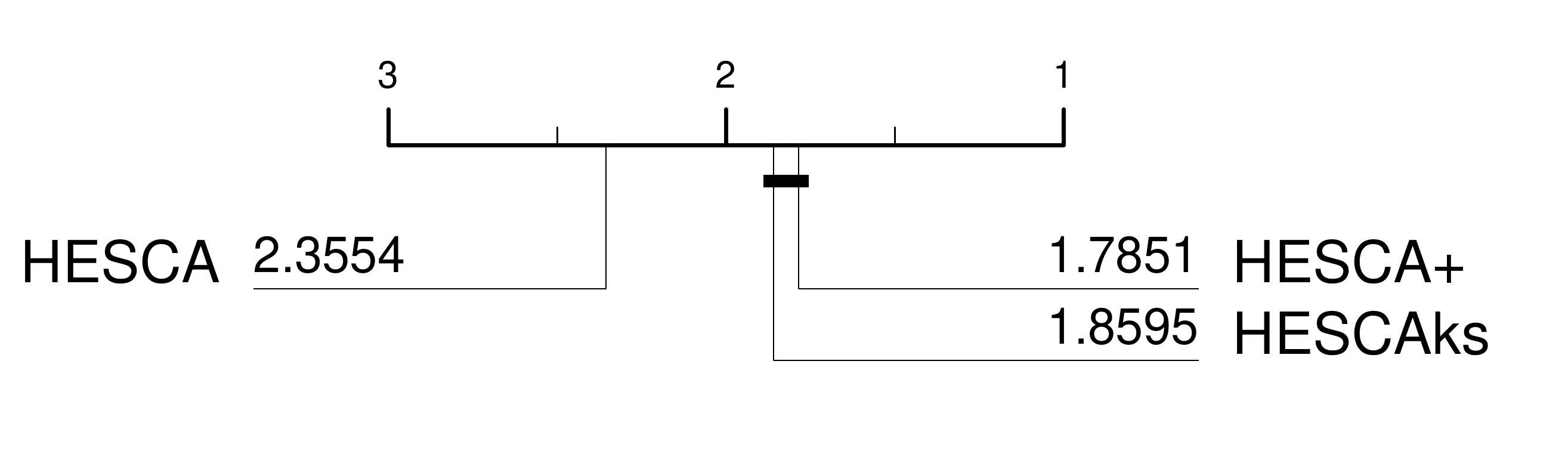}  \\
(a) Error & (b) Balanced Error \\
       \includegraphics[width =6cm, trim={3cm 9cm 0cm 4cm},clip]{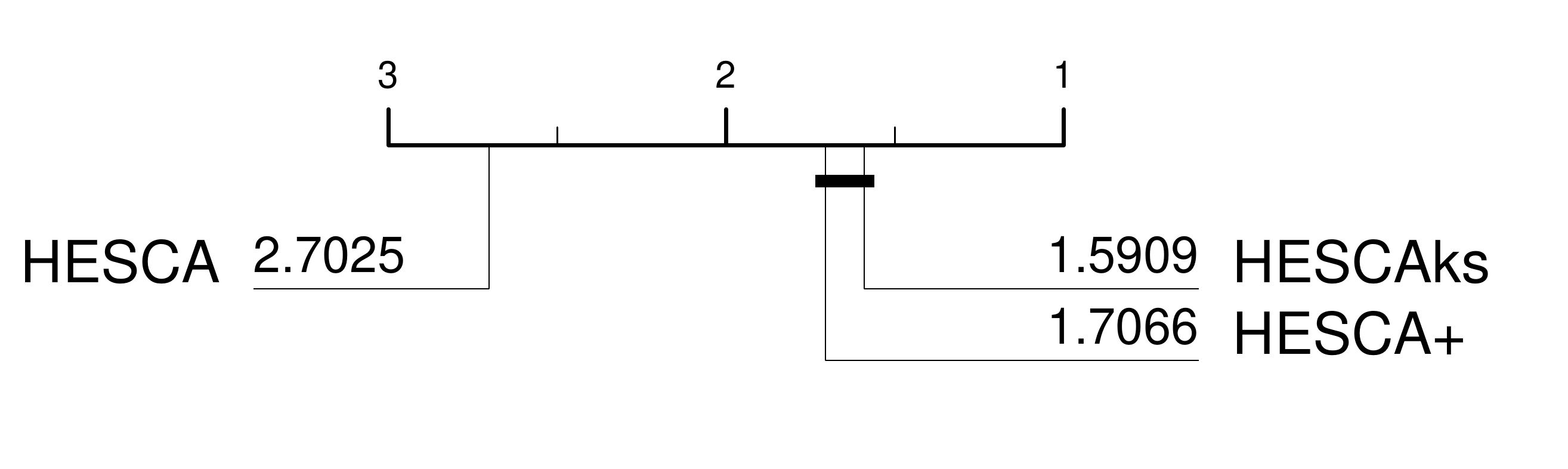}
&
       \includegraphics[width =6cm, trim={3cm 9cm 0cm 4cm},clip]{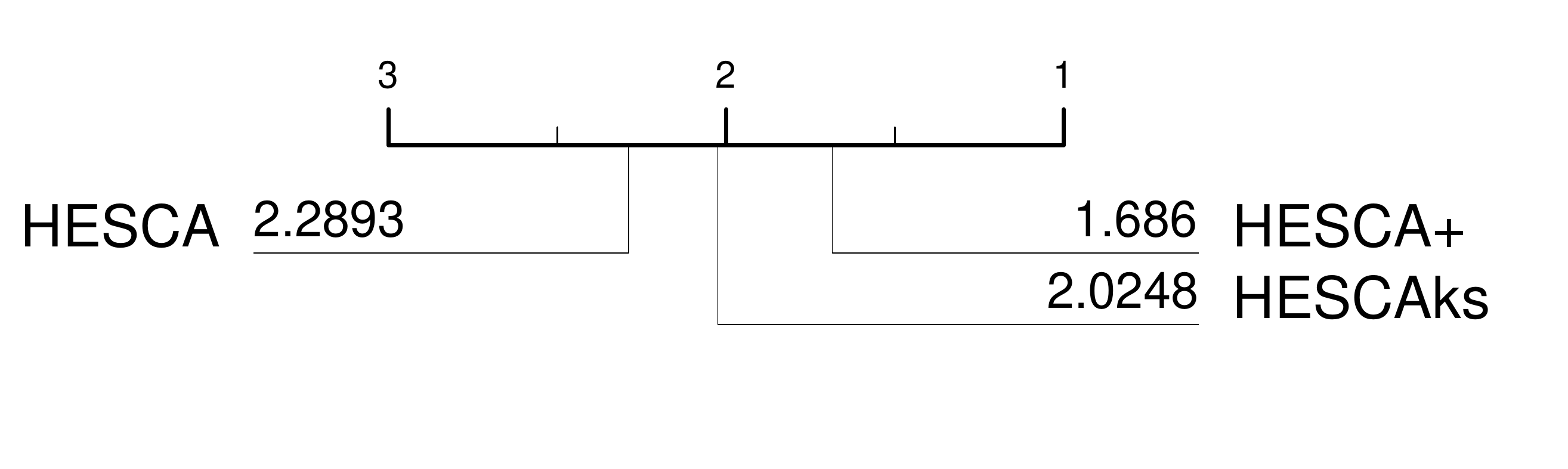}              	\\
(c) AUROC & (b) NLL \\
       \end{tabular}
       \caption{Critical difference diagrams for HESCA, HESCA+ and the kitchen sink version, HESCA$_{ks}$.}
       \label{hesca+ks}
\end{figure}

Finally, given we have the results, we could not resist building an ensemble of all of them, which we call the kitchen sink HESCA (HESCA$_{ks}$). HESCA$_{ks}$ is significantly better than all of its constituents and HESCA. A comparison to HESCA and HESCA+ is shown in Figure~\ref{hesca+ks}. HESCA$_{ks}$ has significantly lower error than HESCA+, there is no difference in AUROC and balanced error and HESCA+ is significantly better in terms of NLL. Adding all these classifiers to HESCA+  brings a small (0.003), but significant, decrease in average error, but it produces significantly worse probability estimates. NLL heavily penalises classifiers when the true class has a very low probability estimate. This indicates that HESCA$_{ks}$ predicts well, but when it gets a case wrong, it tends to get it very wrong (in terms of probability estimate).

\section{Analysis}
\label{analysis}

Comparing overall performance of classifiers is obviously desirable; it addresses the general question: given no other information, what classifier should I use? However, we do have further information. We know the number of train cases, the number of attributes and the number of classes. We also can derive estimates of the error on unseen data from the train data. Does any of this information indicate scenarios where HESCA is gaining an advantage? In Figure~\ref{pickBestUCI} we showed that HESCA and HESCA+ are significantly better than picking the best component and in Section~\ref{tuning} we demonstrated that HESCA and HESCA+ are significantly better that tuned SVMRBF. Can we detect a pattern in these results? Do certain data characteristics explain the improvement? The most obvious factor is train set size. Picking the best classifier based on train estimates is likely to be less reliable with small train sets.
\begin{table}
\caption{HESCA vs pick best split by train set size. The three data sets with the same average error have been removed (acute-inflammation, acute-nephritis and breast-cancer-wisc-diag).
}
\label{byTrainSize}
\begin{tabular}{c|c|c|c}\hline
\#Train Cases 	& \#Problems &	\#HESCA WINS	& Mean Error Difference\\ \hline
1-100			& 28		& 	21		& 1.49\% \\
101-500			& 46		&	36		& 0.71\%\\
501-1000		& 12		&	11		& 1.51\%\\
1001-5000		& 23		&	11		&0.16\% \\
$>$5001			& 9			&	2		& 0.02\%\\ \hline
\end{tabular}
\end{table}


Table~\ref{byTrainSize} breaks down the results given in Figure~\ref{pickBestUCI} by train set size. With under 1000 train cases, HESCA is clearly superior. With 1000-5000 cases, there is little difference. With over 5000 cases, HESCA is better on just 2 of 9 problems, but there is only a tiny difference in error. This would indicate that if one has over 5000 cases then there may be little benefit in using HESCA, although it is unlikely to be detrimental and leads to better estimates of the error on unseen cases. Analysis shows there is no detectable significant effect of number of attributes. For the number of classes, there is a benefit for HESCA on problems with more than 5 classes. HESCA win on 62\% of problems with five or fewer classes (53 out of 85) and wins on 85\% of problems with 6 or more (28 out of 33). This is not unexpected, as a large number of classes is likely to introduce more noise into the estimate of error. This is not caused by deciding on error: we observe the same trend if we choose on balanced error, NLL or AUROC.  There is a similar pattern of results for HESCA+ against pick best, although HESCA+  does better on the problems with over 5000 train cases, winning 4 out of 9.

Some of the problems in this UCI set of data are trivial, in that most classifiers get error less than 5\%. Given we assess classifiers primarily by rank, the gain from HESCA could come from a tiny improvement on these data, where a misclassification on a single case may be the difference between winning and losing. In fact, the opposite is true. On problems where the pick best gets more than 5\% test error, HESCA wins on 76\% (73 out of 96), whereas pick best wins on 14 of the 22 easy problems (although the mean difference is less than 0.5\%). HESCA+ similarly does better on the harder problems.

\begin{figure}[!ht]

       \includegraphics[width =12cm, trim={2cm 3cm 2cm 2.7cm},clip]{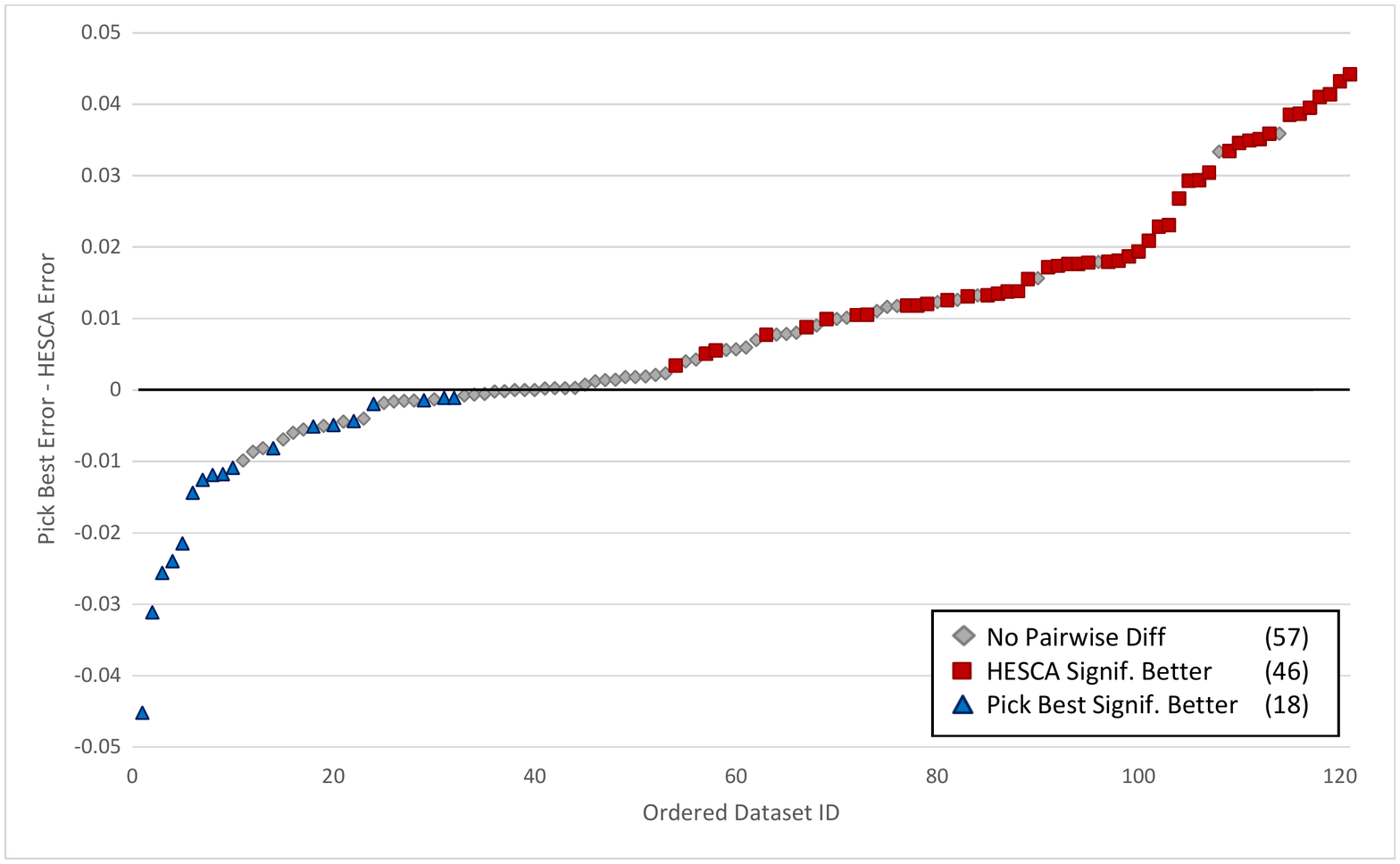}
       \caption{The difference between average errors in sorted order between HESCA and picking the best classifier each time. Significant differences according to paired t-tests over folds are also reported. HESCA is significantly more accurate on 46, the best individual classifier on 18, and there is no significant difference on 57.}
       \label{rankPlot}
\end{figure}

Despite using the same classification algorithms, not all of the differences between pick best and HESCA are small in magnitude. Figure~\ref{rankPlot} shows the ordered differences between the two approaches. The largest difference in favour of HESCA (averaged over 30 folds) is 4.42\% (on the arrhythmia data set) and in favour of pick best 4.5\% (on energy-y1). This demonstrates the importance of the selection method for classifiers; it can cause large differences on unseen data.
This analysis indicates that HESCA is likely to be better approach than simply picking the best when there is not a large amount of training data, there are a large number of classes and/or the problem is hard. Overall, given pick best requires exactly the same amount of work as HESCA, we would recommend using HESCA or HESCA+.

In Section~\ref{results} we showed that both HESCA and HESCA+ are, on average, significantly more accurate than a tuned SVMRBF. However, generally, we are more interested in performance on a new problem. Can we identify data characteristics where the SVM does particularly well or particularly poorly? Table~\ref{trainSizeSVMvsHESCA} and~\ref{trainSizeSVMvsHESCA+} show the results for TunedSVMRBF, HESCA and HESCA+ categorised by number of training cases.

\begin{table}
\caption{HESCA vs TunedSVMRBF by train set size. Four incomplete (miniboone,chess-krvk,magic,adult) and one tie (acute-inflammation) have been removed. }
\label{trainSizeSVMvsHESCA}
\begin{tabular}{c|c|c|c}\hline
\#Train Cases 	& \#Problems &	\#HESCA WINS	& Mean Error Difference\\ \hline
1-100			& 29		& 	24		& 1.74\% \\
101-500			& 47		&	30		& 0.28\%\\
501-1000		& 12		&	8		& 0.19\%\\
1001-5000		& 23		&	10		&0.14\% \\
$>$5001			& 5			&	1		& -0.74\%\\ \hline
\end{tabular}
\end{table}
\begin{table}
\caption{HESCA+ vs TunedSVMRBF by train set size.}
\label{trainSizeSVMvsHESCA+}
\begin{tabular}{c|c|c|c}\hline
\#Train Cases 	& \#Problems &	\#HESCA+ WINS	& Mean Error Difference\\ \hline
1-100			& 29		& 	23		& 2.02\% \\
101-500			& 47		&	28		& 0.81\%\\
501-1000		& 12		&	8		& 1.12\%\\
1001-5000		& 23		&	16		& 1.06\% \\
$>$5001			& 5			&	4		& 0.69\%\\ \hline
\end{tabular}
\end{table}
We observe that the main benefit of HESCA over TunedSVMRBF is with problems with small train set sizes. HESCA+ is also significantly better with small train set sizes, but  maintains a significant advantage for larger problems. These results suggest that as train set size increases the difference between HESCA+ and TunedSVMRBF decreases. However, there is still a difference, and TunedSVMRBF takes an order of magnitude more time to train than HESCA+. We find no pattern of interest in the breakdown by number of attributes. The split by number of classes is shown in Table~\ref{numClassesSVMvsHESCA+}. The proportion of wins for HESCA+ is fairly consistent, but the difference in accuracy is lower for 2-class problems than for those with more than two classes. This may indicate that SVM are better suited to two class problems.

\begin{table}
\caption{HESCA+ vs TunedSVMRBF by number of classes.}
\label{numClassesSVMvsHESCA+}
\begin{tabular}{c|c|c|c}\hline
\#Classes 	& \#Problems &	\#HESCA+ WINS	& Mean Error Difference\\ \hline
2			& 50		& 	32		& 0.65\% \\
3-5			& 34		&	24		& 1.29\%\\
6-10		& 22		&	18		& 2.13\%\\
11+ 		& 10		&	5		& 1.46\% \\ \hline
\end{tabular}
\end{table}

The characteristics of the data can give some general guidance, but ultimately a practitioner is interested in the question of which classifier to use. One way to choose would be based on the estimate of the error/accuracy on the train data. We have already shown that this does not help with the constituents of HESCA, but perhaps it would help choose between HESCA+ and TunedSVMRBF? The problem with the estimate from TunedSVMRBF is that, unless we introduce another level of cross validation, the error on the train data is likely to be biased. One mechanism for assessing how useful the train estimates are is to use a Texas sharpshooter plot (first described in~\cite{batista14cid}).
\begin{figure}[!ht]
	\centering
       \includegraphics[width =10cm, trim={1cm 9.5cm 1cm 1cm},clip]{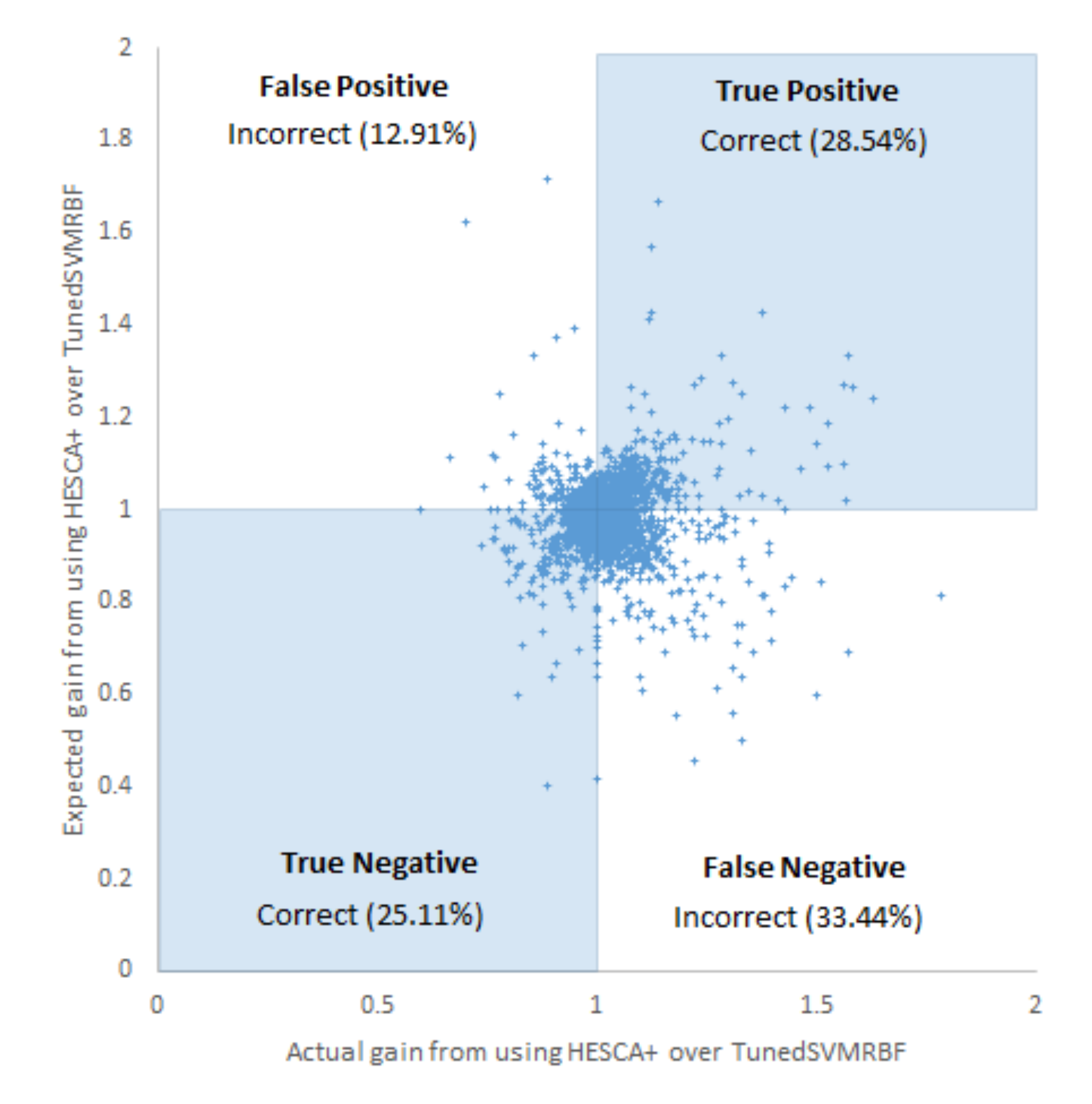}
       \caption{Texas sharp shooter plot for TunedSVMRBF against HESCA+. The top right quadrant contains the problems where both the train and test accuracy for HESCA+ is higher than TunedSVMRBF.}
       \label{texas}
\end{figure}
The basic principle is that the ratio of the training accuracy of two classifiers (generated through cross validation) should give an indication to the outcome for the test data. However, if the cross validation accuracy is biased or subject to high variance, then often the ratio will be misleading. The plot of training accuracy ratio vs. testing accuracy ratio gives a continuous form of contingency table for assessing the usefulness of the training accuracy. If the ratio on training data and testing data are both greater than one then the case is true positive (we predict a gain for one algorithm based on the training data and also observe a gain on the test data); if both ratios are less than one, the problem is a true negative (we predict a loss and also observe a loss). Otherwise, we have an undesirable outcome. If the data sets are evenly spread between the four quadrants, then Batista {\em et al.} observe that we have a situation analogous to the Texas sharpshooter fallacy
(which comes from a joke about a Texan who fires shots at the side of a barn, then paints a target centered on the biggest cluster of hits and claims to be a sharpshooter). Figure~\ref{texas} shows the Texas sharpshooter plot for HESCA+ and TunedSVMRBF, where for the purposes of this graph we deem HESCA+ as being a positive outcome, over all folds and datasets without ties (3361 results). The plot is not too far away from an even spread between the quadarants. The highest proportion of outcomes is False Negative, demonstrating the over optimistic predictions given by TunedSVMRBF due to the parameter optimisation. The decision rule between the classifiers has low sensitivity (0.46) but slightly better specificity (0.66). It may be that a further level of cross validation would improve the accuracy estimates for TuneSVMRBF, but given it is already an order of magnitude slower than HESCA+ and has significantly higher error on average, we would question whether it is worth the effort.

There is a further benefit of having a strong baseline classifier such as HESCA or HESCA+ to compare against; it can provide insights into whether particular parameter settings of another classifier may indicate it will be superior. We performed parameter searches for SVMRBF on 3630 data folds, and on each of these we evaluated 1089 combinations of $C$ and $\gamma$ value. The question we ask is, does SVMRBF tend to do better with any particular range of parameter values? This is a speculative meta-analysis meant more as a test of concept than a definitive contribution to SVM research. With that caveat, we observe the following interesting trends. High values of $\gamma$ ($2^9$ or higher) are rarely selected (71 times on 14 different data sets) but when they are, SVM tends to win (50 out of 71). This indicates that the highly sensitive kernel produced by a large gamma value (often associated with overfitting) is actually detecting discriminatory features smaller $\gamma$ values smooth over. For low values of $\gamma$ ($2^{-16}$ to $2^{-12}$), where the kernel is smooth and SVM becomes an approximation of a linear model, there is little difference between SVM, HESCA and HESCA+, as one would expect, as these would generally be the easiest problems to solve. SVM also does well when the parameter $C$ is in the range -5 to -2, winning 133 out of 214 folds in this region.

To summarise, we observe that the benefit of HESCA and HESCA+ is generally greater with smaller train set sizes and for problems with multiple classes, and that selecting a classifier based on estimates from the train data instead of ensembling will often lead to inferior performance on unseen data. HESCA and HESCA+ provide a strong baseline for any new classification algorithm and comparisons to these may provide insights into the complexity of the problem and the scenarios when an alternative classifier may do better.


\section{Alternative Data Sets: UEA-UCR Time Series Classification Archive}
\label{TSC}
Our interest in heterogeneous ensembles originated in time series classification (TSC) problems, where we ensemble over different representations of the data in a style similar to HESCA~\cite{lines16hive}. TSC involves problems where the attributes are ordered (not necessarily in time) and all real valued. The UCR-UEA repository for TSC contains problems from a wide range of domains such as  classifying image outlines, EEG and spectrographs. There are currently 85 data sets, with diverse data characteristics.

Traditionally, dynamic time warping distance (with window size set through cross validation)~\cite{ratanamahatana05threemyths} with a 1-nearest neighbour classifier (referred to as just DTW henceforth) has been considered the benchmark algorithm for this type of problem. In recent years, a range of bespoke algorithms have been proposed in high impact journals and conferences.
The experimental evaluation in~\cite{bagnall16bakeoff} found that of 18 such algorithms, only 13 were significantly better (in terms of accuracy) than DTW. The best performing algorithm, significantly more accurate than all the others, was the Collective of Transformation-based Ensembles (COTE)~\cite{bagnall15cote}. COTE uses an ensemble structure that is the progenitor of HESCA. It has components built on different representations of the data. Our goal is to test whether our core hypothesis that HESCA significantly improves basic components on data independent of the UCI data, then to examine how well it performs in comparison to bespoke algorithms designed specifically for time series. To do so, we ignore the ordering of the series and treat each time step in the series as a feature for traditional vector based classification. The UCR-UEA data sets generally have many more features than the UCI data. This has meant we have had to make one change to HESCA: we remove logistic regression because it cannot feasibly be built on many of the data. This is the only change that has been made, HESCA and HESCA+ have otherwise not been altered to run on the time series data. Since DTW is a 1-nearest neighbour classifier, it always produces 0/1 probability estimates. Because of this, we omit a probabilistic evaluation, as it has little meaning for DTW.

\begin{figure}[!ht]
	\centering
\begin{tabular}{cc}
       \includegraphics[width =6cm, trim={4cm 7cm 1.5cm 3cm},clip]{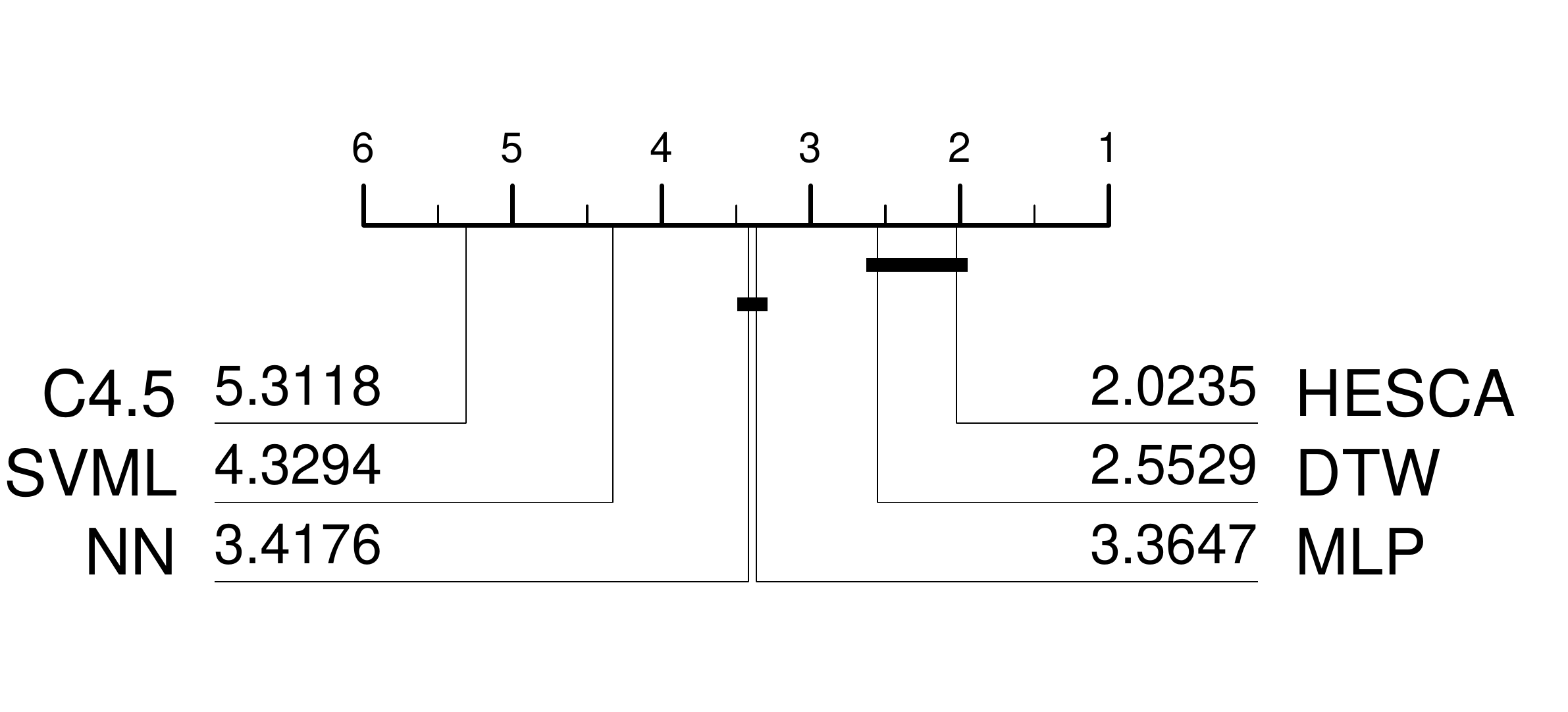}              	
&
       \includegraphics[width =6cm, trim={4cm 7cm 1.5cm 3cm},clip]{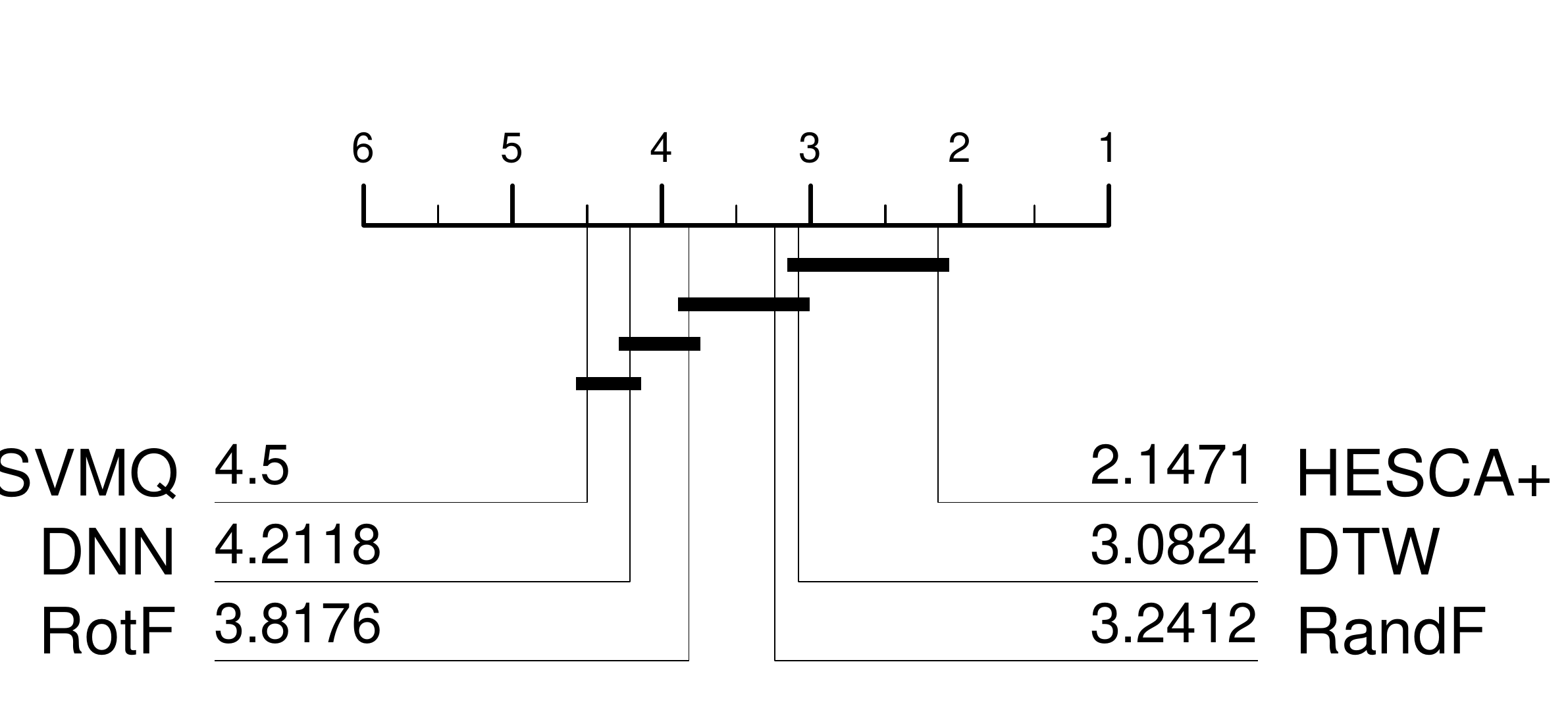}  \\
       (a) & (b)
       \end{tabular}
       \caption{Average ranked errors for (a) HESCA and (b) HESCA+ against their components and DTW on the 85 datasets in the UCR-UEA archive.}
       \label{ucrHesca}
\end{figure}

Figure~\ref{ucrHesca} shows the critical difference diagrams for accuracy of HESCA, its constituents and DTW. Both HESCA and HESCA+ are significantly better than their components. These results closely mirror those on the UCI datasets presented in Figures~\ref{hescaUCI} and~\ref{hesca+UCI}. Furthermore, neither HESCA nor HESCA+ are significantly worse than DTW. This should be considered in the context that neither classifier takes advantage of any information in the ordering of attributes. HESCA+ has a higher average rank than 9 of the 18 bespoke time series classification algorithms evaluated in~\cite{bagnall15cote}, and is not significantly worse than 11 of them. HESCA+, a simple classifier using off the shelf components and a simple weighting scheme, is as accurate as complex algorithms that use a range of complicated techniques such as forming bags of patterns, using edit distance based similarity, differential based distances, compression techniques and decision trees based on short subseries features.

\section{Conclusions}
\label{conc}
The key message of this paper is simple: forming heterogeneous ensembles of approximately equivalent classifiers  produces a significantly better classifier (in terms of error, ordering and probability estimates) than a wide range of potential  base classifiers,  homogeneous ensembles or a tuned support vector machine using an RBF kernel. The HESCA ensembling scheme uses the very simple method of weighting based on estimates of the error formed on the train data, and we found this technique as good as, or better than, more complex combination schemes such as Confusion Entropy. We have supported these claims with a wide range of experiments on data with real valued attributes from two distinct data archives. For a particular problem, we would recommend the use of HESCA in the first instance. This will at the very least give a sound benchmark to which to compare other algorithms, but it is also likely to produce a solution as good or better than more complex algorithms with orders of magnitude less computation. HESCA+ is slower to build, but on average is significantly better than classifiers many would consider state-of-the-art. We think it is particularly beneficial to use HESCA or HESCA+ on problems with fewer than 1000 training cases and more than two classes. We have provided a Weka based HESCA implementation that is flexible and can be used directly with the methods in the  {\texttt Classifier} interface. It can be tailored to include any classifiers available in Weka or to load results from file for classifiers built in an alternative tool kit. We have released all our data, code and results, to facilitate the further study of algorithm performance. All our experiments can be reproduced on the exact training testing folds used. We envisage both HESCA and HESCA+ as solid benchmarks rather than state-of-the-art classifiers; it may well be possible that advanced classifiers such as deep learning and support vector machines can be designed to beat them both, but if this is the case it is not trivial. Ultimately we hope to drive a better understanding of what classifier to use for a new problem and how best to use it. However, with current technology, our conclusion is that, rather than expend extra computational time tuning a single classifier, it is better ensemble different classifiers from different families of algorithms.

\section*{Acknowledgements}
This work is supported by the UK Engineering and Physical Sciences Research Council (EPSRC)  [grant number EP/M015087/1]and the Biotechnology and Biological Sciences Research Council [grant number BB/M011216/1]. The experiments were carried out on the High Performance Computing Cluster supported by the Research and Specialist Computing Support service at the University of East Anglia and using a Titan X Pascal donated by the NVIDIA Corporation.

\bibliographystyle{plain}


\end{document}